\documentclass[letterpaper, 11pt]{article}

\usepackage[utf8]{inputenc}
\usepackage{import}
\usepackage[margin=1in]{geometry}
\usepackage{setspace}\onehalfspace
\usepackage{microtype}
\usepackage{adjustbox}
\usepackage{multirow}
\usepackage{graphicx}
\usepackage[]{subcaption}
\usepackage{algorithm}
\usepackage{algpseudocode}
\usepackage{longtable}
\usepackage{pdflscape}
\usepackage{rotating, tabularx}

\usepackage{amsmath}
\usepackage{amssymb}
\usepackage{bm}
\usepackage{natbib}
\usepackage{amsthm}

\usepackage{booktabs} %
\usepackage[hyphens]{url}
\usepackage[bookmarks=false,hidelinks]{hyperref}

\usepackage{mathtools}
\usepackage{color}
\usepackage[table]{xcolor}
\usepackage{authblk}

\usepackage{tikz}
\usetikzlibrary{matrix,backgrounds}
\usetikzlibrary{decorations.pathreplacing,calligraphy}

\usetikzlibrary{positioning}

\tikzstyle{chromosome} = [
	matrix of math nodes, ampersand replacement=\&,
	nodes={draw, anchor=center, minimum width=2.5em},
	nodes in empty cells, minimum height=1.8em
]

\tikzstyle{tour} = [
	matrix of math nodes, ampersand replacement=\&,
	nodes={draw, anchor=center, minimum width=1.5em},
	nodes in empty cells, minimum height=1em
]

\tikzstyle{depotnode} = [shape=circle, draw, fill=white!20!black, text=white]
\tikzstyle{customernode} = [shape=circle, draw, fill=white!40!blue, text=white]

\newcommand{\email}[1]{{\href{mailto:#1}{\nolinkurl{#1}}}}

\usepackage{bm}

\newcommand{\set}[1]{\mathcal{#1}}

\usepackage{xcolor}

\usepackage{etoolbox}
\makeatletter
\def\do#1{\@namedef{#1c}{\ensuremath{\mathcal{#1}}}}
\docsvlist{A, B, C, D, E, F, G, H, I, J, K, L, M, N, O, P, Q, R, 
S, T, U, V, W, X, Y, Z}
\makeatother

\usepackage{bbm}

\newcommand{\SPLIT}{\textsc{Split}}
\newcommand{\JOIN}{\textsc{Join}}

\algnewcommand\algorithmicforeach{\textbf{for each}}
\algdef{S}[FOR]{ForEach}[1]{\algorithmicforeach\ #1\ \algorithmicdo}

\title{A Hybrid Genetic Algorithm for the min-max Multiple Traveling Salesman Problem}

\author{Sasan Mahmoudinazlou}
\affil{Department of Industrial and Management Systems Engineering, University of South Florida, Tampa, FL 33620, U.S.A.\\\email{sasanm@usf.edu}}
\author{Changhyun Kwon\thanks{Corresponding author}}
\affil{Department of Industrial and Systems Engineering, KAIST, Daejeon, 34141, Republic of Korea\\\email{chkwon@kaist.ac.kr}}

\date{}

\begin{document}

\maketitle

\begin{abstract}
This paper proposes a hybrid genetic algorithm for solving the Multiple Traveling Salesman Problem (mTSP) to minimize the length of the longest tour. The genetic algorithm utilizes a TSP sequence as the representation of each individual, and a dynamic programming algorithm is employed to evaluate the individual and find the optimal mTSP solution for the given sequence of cities. A novel crossover operator is designed to combine similar tours from two parents and offers great diversity for the population. For some of the generated offspring, we detect and remove intersections between tours to obtain a solution with no intersections. This is particularly useful for the min-max mTSP. The generated offspring are also improved by a self-adaptive random local search and a thorough neighborhood search. Our algorithm outperforms all existing algorithms on average, with similar cutoff time thresholds, when tested against multiple benchmark sets found in the literature. Additionally, we improve the best-known solutions for $21$ out of $89$ instances on four benchmark sets. 
\end{abstract}

\section{Introduction}
The Traveling Salesman Problem (TSP) is a widely recognized combinatorial optimization problem.
This problem determines the minimal-cost Hamiltonian cycle---one that visits every city exactly once---using a given set of cities and their respective distances. 
There are numerous applications for the TSP in the fields of transportation, logistics, and manufacturing. 

This paper studies the Multiple Traveling Salesman Problem (mTSP).
As a variant of the TSP, an mTSP includes multiple salesmen who start and end their tours at the same depot.
The following graph notation is used to represent the problem.
We consider $\set{G} = (\set{V}, \set{E})$, a complete undirected graph with $n$ vertices representing the cities to be visited.
We let $m$ be the number of salesmen available for the task and $d$ be the depot vertex from where each salesman starts and ends their tour.
We denote the distance between vertex $i$ and vertex $j$ by $c_{ij}$.

The purpose of the mTSP is to find a set of $m$ tours, one for each salesman, such that every city is visited exactly once by one of the salesmen. 
There are two common objective functions for solving the mTSP. 
The first objective involves minimizing the overall distance traveled by all the salesmen, a problem known as the min-sum mTSP \citep{lin1973effective}.
The alternative objective is to minimize the total distance covered by the salesperson with the longest tour, which is commonly referred to as the min-max objective function \citep{francca1995m}
In our paper, we focus on solving the min-max mTSP problem.

The mTSP solution can be represented using a set of $m$ separate TSP tours, one for each salesman, starting and ending at the depot vertex $d$. 
In other words, we need to partition the set of vertices $\set{V}$ into $m$ disjoint subsets $\set{V}_1, \set{V}_2, ..., \set{V}_m$, where each subset represents the cities visited by a single salesman. 
Each salesman's tour can be represented as a sequence of vertices that form a Hamiltonian cycle within the corresponding subset of cities.
The overall solution can then be represented as a concatenation of the $m$ Hamiltonian cycles in the order in which the salesmen visit the cities.

Note that mTSPs have similar applications as TSPs and are introduced in scenarios in which multiple agents need to cover a set of locations concurrently. 
The mTSP can, for example, be used to optimize delivery routes for multiple vehicles or drivers in logistics and transportation. 
mTSP can be used in network design to optimize the placement of facilities such as communication towers and data centers. 
In manufacturing, the mTSP can be used to optimize the routing of multiple machines or robots that need to work together to produce a product.

The mTSP belongs to the NP-hard class of problems, meaning that its exact solution is computationally impractical when dealing with large data sets. 
It is, therefore, necessary to use heuristic algorithms in order to obtain near-optimal solutions within a reasonable period of time. 
One popular approach is to use genetic algorithms (GA), which are stochastic optimization algorithms inspired by the principles of natural selection and genetics.

In this paper, we propose a hybrid genetic algorithm for solving the mTSP. 
Our algorithm employs a divide-and-conquer approach, where each individual in the genetic algorithm represents a TSP solution. 
To determine the optimal mTSP solution for a given TSP sequence, we utilize an algorithm based on dynamic programming.
To enhance the diversity of the population, we have developed a new crossover function. 
This function promotes the exchange of genetic information between similar tours from two parents, thereby introducing greater variation among the individuals.
For refining the solutions, we employ a combination of local search methods. 
These methods aim to further optimize the tours by exploring the neighborhood of each solution.
Additionally, we introduce a function specifically designed to detect and remove the intersections between tours. 
This function ensures that the generated solutions have no intersections, which is particularly beneficial for achieving high-quality solutions in the mTSP.
Nevertheless, it is not applied to every individual in the population since there can exist intersections within an optimal solution of the mTSP.

Our paper makes four significant contributions. 
Firstly, we utilize a method that combines dynamic programming and genetic algorithms to tackle the mTSP. 
This approach leverages the strengths of both techniques, aiming to improve solutions in the context of this problem.
Secondly, we introduce a new crossover function that offers two advantages: it is computationally inexpensive and highly effective in diversifying populations. 
This contributes to the exploration of a wider solution space and enhances the algorithm's performance.
Thirdly, we show that the removal of intersections between tours can lead to better solutions in min-max mTSP. 
Lastly, we validate the effectiveness of our algorithm through extensive experiments on benchmark instances. 
Our results not only surpass those of existing algorithms but also demonstrate notable improvements in the best solutions found within a specified time period for multiple instances.
These contributions collectively showcase the value and efficacy of our proposed approach for solving the mTSP.

The remainder of this paper is organized as follows. 
In Section \ref{sec:literature}, we provide a literature review of existing methods for solving the mTSP, highlighting their strengths and limitations. 
Section \ref{sec:method} provides a detailed description of our hybrid genetic algorithm, explaining the key components and their interplay in addressing the mTSP.
In Section \ref{sec:results}, we present the results of our experiments on benchmark instances. 
Finally, in Section \ref{sec:conclusion}, we summarize our findings and contributions, emphasizing the significance of our approach to solving the mTSP.
We also discuss potential directions for future research, exploring avenues for further improvement and extension of our algorithm.

\section{Literature review} \label{sec:literature}

The multiple traveling salesman problem (mTSP) has garnered significant attention in the literature, resulting in extensive research efforts. 
Among the various approaches proposed, the Genetic Algorithm (GA) has emerged as one of the most popular and successful methods for solving the mTSP. 
In this section, we review the existing literature that focuses on the utilization of GA for tackling the mTSP, emphasizing its efficiency and effectiveness in addressing the problem.
Subsequently, we shift our discussion to explore other heuristics employed in the solution of the mTSP. 
Lastly, we delve into the variants of the mTSP and discuss the relevant literature dedicated to their study. 
Notably, many of these studies also employ evolutionary algorithms as a primary tool for addressing the specific problem variants.
By providing a comprehensive review of the literature in these three categories, we aim to present a holistic understanding of the research landscape surrounding the mTSP and its various solution methodologies.

\begin{figure}
    \begin{subfigure}[b]{\linewidth}\centering
        \resizebox{4.5in}{!}{%
            \begin{tikzpicture}
                \matrix[chromosome] (P1) 
                {
                   4 \& 
                   1 \& 
                   3 \&
                   5 \&
                   -1 \&
                   6 \&
                   2 \&
                   10 \&
                   -1 \&
                   8 \&
                   9 \&
                   7 \\
                };
                \draw[ultra thick, decorate, decoration={calligraphic brace, amplitude=10pt}]
                (P1-1-4.south east) -- (P1-1-1.south west) node[below=1em, midway] {salesperson 1};
                \draw[ultra thick, decorate, decoration={calligraphic brace, amplitude=10pt}]
                (P1-1-8.south east) -- (P1-1-6.south west) node[below=1em, midway] {salesperson 2};
                \draw[ultra thick, decorate, decoration={calligraphic brace, amplitude=10pt}]
                (P1-1-12.south east) -- (P1-1-10.south west) node[below=1em, midway] {salesperson 3};
            \end{tikzpicture}%
            
        }
        \caption{Representation used in \citet{tang2000multiple}. The tours of different salesmen are separated by $-1$.}
        \label{fig:one-chromosome}	
    \end{subfigure}
    
    \par\bigskip %

    \begin{subfigure}[b]{\linewidth}\centering
       
        \resizebox{4in}{!}{%
            \begin{tikzpicture}
                \matrix[chromosome] (P1) 
                {
                   6 \& 
                   4 \& 
                   1 \&
                   8 \&
                   2 \&
                   3 \&
                   9 \&
                   5 \&
                   7 \&
                   10 \\
                };
                \matrix[below = 1em of P1, chromosome] (P2) 
                {
                   2 \& 
                   1 \& 
                   1 \&
                   3 \&
                   2 \&
                   1 \&
                   3 \&
                   1 \&
                   3 \&
                   2 \\
                };
                \node at ([yshift=10pt] P1-1-5.north) {Cities};
                \node at ([yshift=10pt] P2-1-5.north) {Salespersons};
            \end{tikzpicture}%
        }

       \caption{Two chromosome representation. The first chromosome is the sequence of cities, the second is the index of salesmen visiting them.}
       \label{fig:two-chromosome}
    \end{subfigure}
    
    \begin{subfigure}[b]{\linewidth}\centering
        \resizebox{5in}{!}{%
            \begin{tikzpicture}
                \matrix[chromosome] (P1) 
                {
                   4 \& 
                   1 \& 
                   3 \&
                   5 \&
                   6 \&
                   2 \&
                   10 \&
                   8 \&
                   9 \&
                   7 \&
                   4 \&
                   3 \&
                   3 \\
                };
                \draw[ultra thick]
                ([yshift=10pt] P1-1-10.north east) -- ([yshift=-10pt] P1-1-10.south east);
                \node at ([yshift=10pt] P1-1-6.north) {Cities};
                \node at ([yshift=20pt] P1-1-12.north) {Cities per};
                \node at ([yshift=10pt] P1-1-12.north) {Salespersons};
                \draw[ultra thick, decorate, decoration={calligraphic brace, amplitude=10pt}]
                (P1-1-4.south east) -- (P1-1-1.south west) node[below=1em, midway] {Salesperson 1} ;
                \draw[ultra thick, decorate, decoration={calligraphic brace, amplitude=10pt}]
                (P1-1-7.south east) -- (P1-1-5.south west) node[below=1em, midway] {Salesperson 2};
                \draw[ultra thick, decorate, decoration={calligraphic brace, amplitude=10pt}]
                (P1-1-10.south east) -- (P1-1-8.south west) node[below=1em, midway] {Salesperson 3};
                \draw[->, ultra thick] ([xshift=0pt, yshift=-10pt]P1-1-11.south) |- ++(0, -1) |- ++(-8,0)
                   -| ([yshift=-25pt]P1-1-2.south east);
               \draw[->, ultra thick] ([xshift=0pt, yshift=-10pt]P1-1-12.south) |- ++(0, -1.3) |- ++(-5.88,0)
                   -| ([yshift=-25pt]P1-1-6.south);
               \draw[->, ultra thick] ([xshift=0pt, yshift=-10pt]P1-1-13.south) |- ++(0, -1.6) |- ++(-2.88,0)
                   -| ([yshift=-25pt]P1-1-9.south);
            \end{tikzpicture}%
        }
        \caption{Two-part chromosome representation.}
        \label{fig:two-part-chromosome1}	
    \end{subfigure}
    
    \par\bigskip %
    
    \begin{subfigure}[b]{\linewidth}\centering
        \resizebox{5in}{!}{%
            \begin{tikzpicture}
                \matrix[chromosome] (P1) 
                {
                   1.2 \& 
                   2.2 \& 
                   1.3 \&
                   1.1 \&
                   1.4 \&
                   2.1 \&
                   3.3 \&
                   3.1 \&
                   3.2 \&
                   2.3 \\
                };
                \matrix[right = 1em of P1, chromosome] (P2) 
                {
                   1 \& 
                   2 \& 
                   3 \\
                };
                \draw[thick]
                ([xshift=5 pt] P1-1-10.east) -- ([xshift=-5 pt]P2-1-1.west);
            \end{tikzpicture}%
        }
        \caption{GGA chromosome encoding. The value of 3.2 in the gene
    position nine means that salesperson three visits city nine, and it is the
    second city on the route. The second part comprises the group section of the chromosome}
        \label{fig:gga-chromosome}	
    \end{subfigure}
    
    \par\bigskip %
    
    \begin{subfigure}[b]{\linewidth}\centering
        \resizebox{4in}{!}{%
            \begin{tikzpicture}
                \matrix[chromosome] (P1) 
                {
                   4 \& 
                   1 \& 
                   3 \&
                   5 \\
                };
                \matrix[right = 1em of P1, chromosome] (P2) 
                {
                   6 \& 
                   2 \& 
                   10 \\
                };
                \matrix[right = 1em of P2, chromosome] (P3) 
                {
                   8 \& 
                   9 \& 
                   7 \\
                };
                \node[] at ([xshift=-7pt] P1-1-1.west) {\Huge\{};
                \node[] at ([xshift=7pt] P3-1-3.east) {\Huge\}};
                \node[ultra thick] at ([xshift=10pt, yshift = -5pt] P1-1-4.east) {,};
                \node[ultra thick] at ([xshift=10pt, yshift = -5pt] P2-1-3.east) {,};
            \end{tikzpicture}%
        }
        \caption{Many-chromosome representation.}
        \label{fig:many-chromosome}	
    \end{subfigure}
    
    \par\bigskip %
    
    \begin{subfigure}[b]{\linewidth}\centering
        \resizebox{5in}{!}{%
            \begin{tikzpicture}
                \matrix[chromosome] (P1) 
                {
                   4 \& 
                   1 \& 
                   3 \&
                   5 \&
                   6 \&
                   2 \&
                   10 \&
                   8 \&
                   9 \&
                   7 \&
                   4 \&
                   7 \\
                };
                \draw[ultra thick]
                ([yshift=10pt] P1-1-10.north east) -- ([yshift=-10pt] P1-1-10.south east);
                \node at ([yshift=10pt] P1-1-6.north) {Cities};
                \node at ([yshift=10pt] P1-1-12.north) {Break points};
                \draw[ultra thick, decorate, decoration={calligraphic brace, amplitude=10pt}]
                (P1-1-4.south east) -- (P1-1-1.south west) node[below=1em, midway] {Salesperson 1} ;
                \draw[ultra thick, decorate, decoration={calligraphic brace, amplitude=10pt}]
                (P1-1-7.south east) -- (P1-1-5.south west) node[below=1em, midway] {Salesperson 2};
                \draw[ultra thick, decorate, decoration={calligraphic brace, amplitude=10pt}]
                (P1-1-10.south east) -- (P1-1-8.south west) node[below=1em, midway] {Salesperson 3};
                \draw[->, ultra thick] ([xshift=0pt, yshift=-10pt]P1-1-11.south) |- ++(0, -0.7) |- ++(-6,0)
                   -| ([yshift=-5 pt]P1-1-4.south east);
               \draw[->, ultra thick] ([xshift=0pt, yshift=-10pt]P1-1-12.south) |- ++(0, -1) |- ++(-4,0)
                   -| ([yshift=-5pt]P1-1-7.south east);
            \end{tikzpicture}%
        }
        \caption{Two-part chromosome representation with breakpoints.}
        \label{fig:two-part-chromosome2}	
    \end{subfigure}
    
    \vspace{0.5cm}
    
    \begin{subfigure}[b]{\linewidth}\centering
        \resizebox{4in}{!}{%
            \begin{tikzpicture}
                \matrix[chromosome] (P1) 
                {
                   4 \& 
                   1 \& 
                   3 \&
                   6 \&
                   2 \&
                   10 \&
                   8 \&
                   9 \&
                   7 \\
                };
            \end{tikzpicture}%
            
        }
        \caption{Chromosome representation used in this study, which is a sequence of all cities.}
        \label{fig:tsp}	
    \end{subfigure}
    
    \caption{Different chromosome encodings used in the literature.}
    \end{figure}
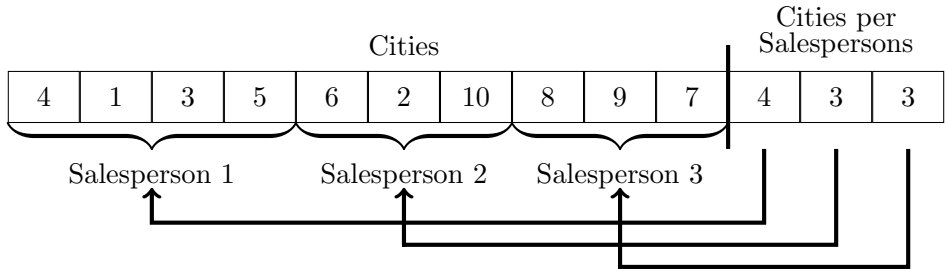
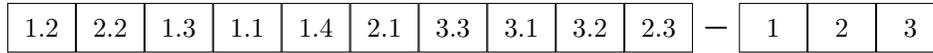
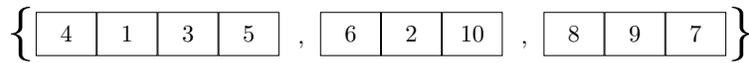
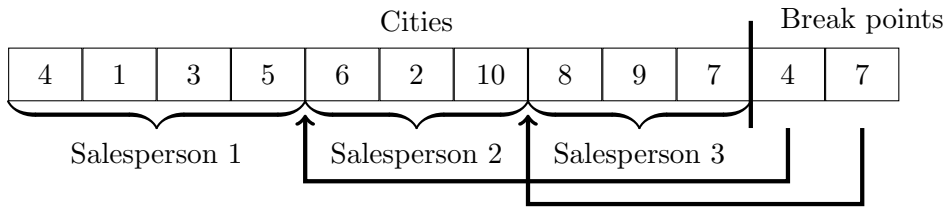
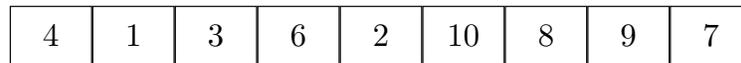

Genetic algorithms have emerged as the predominant approach for solving mTSP problems in the literature. 
\citet{tang2000multiple} address the mTSP as part of a rolling scheduling system and employs a chromosome encoding scheme that includes the sequence of nodes along with delimiters to separate the tours for each salesperson (see Figure \ref{fig:one-chromosome}).
Using two-chromosome encoding, \citet{park2001hybrid} develops a GA to solve the Vehicle Routing Problem with Time Windows (VRPTW). 
The first chromosome represents the cities, and the second represents the salesmen who visit those cities (Figure \ref{fig:two-chromosome}). 
The study incorporates partially mapped crossover (PMX) and explores four different mutation functions.
\citet{carter2006new} propose a genetic algorithm for solving the mTSP using a two-part chromosome encoding. 
The first part represents the sequence of cities, and the second part indicates the number of cities visited by each salesperson (see Figure \ref{fig:two-part-chromosome1}). 
Furthermore, \citet{chen2011operators} conduct a design of experiments to explore different combinations of crossovers and mutations based on the aforementioned two-part chromosome encoding. 
This approach allows for the evaluation of various genetic operators' effectiveness in solving the mTSP.

In recent years, several studies have aimed to enhance genetic algorithms for solving the mTSP through the introduction of new crossovers or novel chromosome encodings. 
Notably, \citet{yuan2013new} propose a new crossover named TCX, specifically tailored for the two-part chromosome representation. 
TCX operates by treating each salesman separately during the crossover process in the first part of the chromosome, ultimately improving the effectiveness of the algorithm.
An alternative approach to chromosome encoding is explored by \citet{brown2007grouping}, which introduces the Grouping Genetic Algorithm (GGA) chromosome representation for encoding mTSP solutions (see Figure \ref{fig:gga-chromosome}). 
In this encoding, each city carries information about the assigned salesman and its order within the salesman's tour.
\citet{singh2009new} develop a grouping genetic algorithm using a many-chromosome representation, where each part of a chromosome represents a tour for a specific salesman (see Figure \ref{fig:many-chromosome}). 
They also devise a new crossover strategy: in each iteration, the parent is randomly selected, and the shortest tour from the selected parent is added to the offspring. 
The cities belonging to this tour are removed from the other parent. 
This process is repeated $m-1$ times, followed by the addition of the remaining cities based on either a greedy or random approach. 
This study served as the initial inspiration for our own crossover development.
Furthermore, \citet{wang2017memetic} develop a Memetic Algorithm with a many-chromosome representation utilizing Sequential Variable Neighborhood Descent (MASVND) to address the min-max mTSP.
By incorporating these innovative approaches, these studies have contributed to the ongoing advancement and improvement of genetic algorithms for solving the mTSP.

This study proposes a hybrid approach that combines a genetic algorithm, dynamic programming, and local searches to solve the mTSP. 
The motivation for using this approach stems from our previous work \citep{mahmoudinazlou2023hybrid}, where a divide and conquer approach is employed to solve the Traveling Salesman Problem with Drones (TSPD). 
In that study, a GA with chromosome representations incorporating the sequences of the truck and drone is used. 
By utilizing a dynamic programming-based algorithm called \JOIN{}, the optimal location for the drone's launch and landing is determined.
Building upon this previous work, the current study adopts a similar approach by utilizing a GA with a simple chromosome representation consisting of TSP sequences (see Figure \ref{fig:tsp}). 
To obtain the mTSP solution, a dynamic programming algorithm called \SPLIT{} is employed to determine the optimal delimiters for the given TSP sequence. 
This approach transfers some decision-making from the GA to the \SPLIT{} algorithm, thereby enhancing decision-making effectiveness and reducing the complexity of the GA. 
This simplification allows the GA to reach convergence more quickly. 
Furthermore, this approach enables simultaneous searches in two different regions. 
The GA focuses on exploring the TSP neighborhood, while the local search functions aim to improve the mTSP solution within the mTSP region.
By combining GA, dynamic programming, and local searches, this hybrid approach offers several advantages compared to existing methods, including improved convergence speed, optimal decision-making through dynamic programming, and simultaneous searches in multiple areas to enhance the overall solution quality.

Several alternative heuristics have been proposed for solving the mTSP. 
These methods offer different perspectives and strategies to tackle the problem.
\citet{junjie2006ant} apply an Ant Colony Optimization (ACO) algorithm to solve the min-sum mTSP. 
Similarly, \citet{liu2009ant} develop an Ant Colony system to address mTSP with multiple objectives.
\citet{venkatesh2015two} introduce two metaheuristic algorithms: one based on the Artificial Bee Colony (ABC) algorithm and another based on the Invasive Weed Optimization (IWO) algorithm. 
These algorithms are combined with a local search procedure that involves a 2-opt algorithm and a heuristic for reallocating cities from the longest tour to other tours in a greedy manner.
\citet{soylu2015general} proposes a General Variable Neighborhood Search (GVNS) for the mTSP. 
The study utilizes six different local search functions, including one-point move, two-point move, Or-opt2 move, Or-opt3 move, three-point move, and 2-opt move, to explore the search space effectively.
\citet{he2022hybrid} introduce a Hybrid Search with Neighborhood Reduction (HSNR) algorithm for solving the mTSP. 
The HSNR algorithm follows a two-step iterative process, involving inter-tour and intra-tour searches, to achieve convergence. 
In the inter-tour step, two neighborhood search functions, an insert operator, and a cross-exchange are employed based on Tabu Search to explore the search space between the tours. 
The intra-tour optimization is performed using the TSP heuristic EAX, which focuses on improving the individual tours.
\citet{zheng2022effective} propose an iterated two-stage heuristic algorithm (ITSHA), which includes initialization as stage one and improvement as stage two.
In the first stage, the authors create initial solutions by either using a random greedy process or employing C-means clustering.  
The improvement stage involves a variable neighborhood search that includes a 2-opt move, an insert move, and a swap move.

To the best of our knowledge, \citet{zheng2022effective}'s mTSP results are the best on widely used benchmark sets at the time of writing this paper.
However, it is worth noting that \citet{he2023memetic}'s Memetic Algorithm (MA) has outperformed their algorithm with a longer cutoff time.
Nevertheless, there are no experimental results available for their MA using the commonly used cutoff time.
Therefore, in section \ref{sec:results}, we compare our results primarily with the ITSHA algorithm.

A number of valuable studies have been conducted that utilize reinforcement learning to solve the mTSP. 
The following are some of these papers that we would like to mention. 
\citet{hu2020reinforcement} propose a multi-agent reinforcement learning algorithm that utilizes GNNs as the training tool to solve mTSP. 
A multi-agent reinforcement learning algorithm, ScheduleNet, is proposed by \citet{parklearn} for solving the min-max mTSP. 
\citet{kim2022neuro} employ GNNs in order to train an agent to use the cross-exchange neighborhood for the solution of the mTSP. 
Their algorithm is known as the Neuro Cross Exchange (NCE) algorithm, whose results on some benchmark sets appeared promising. 
Therefore, we chose NCE as one of the baseline algorithms for evaluating our method. 

The mTSP family contains a number of variants that deserve attention. 
\citet{alves2015using} propose a bi-objective mTSP that attempts to minimize both the total distance traveled and the longest tour. 
There are two approaches presented, one that consists of a multi-objective GA and one that consists of a regular GA that combines two objectives. 
This study utilizes the two-chromosome encoding (Figure \ref{fig:two-chromosome}). 
As a result of their findings, multi-objective GA performs better for this bi-objective mTSP.
In a variant of the mTSP introduced by \citet{li2013new}, certain cities are restricted to a particular salesperson. 
This problem is referred to as mTSP*. 
The problem is also solved using GA with two-chromosome encoding. 
A similar problem is studied by \citet{liu2021comparative} under the name Visiting Constrained Multiple Traveling Salesman Problem (VCMTSP). 
According to this problem, each city is restricted to only being visited by certain salesmen who have been predetermined. 
An evolutionary algorithm with many-chromosome encoding (Figure \ref{fig:many-chromosome}) and five local search functions is proposed in this study.

There is also a variant of mTSP where there are multiple depots. 
This problem is known as multi-depot mTSP or M-mTSP. 
A review of some of the most recent M-mTSP studies is presented. 
A new ant colony optimization algorithm is proposed by \citet{lu2019mission} for solving M-mTSP. 
\citet{jiang2020new} propose a hybrid Partheno Genetic Algorithm (PGA) and an ant colony optimization algorithm to solve the M-mTSP. 
The solution is encoded using a two-part chromosome representation, with the second part representing the breakpoints (Figure \ref{fig:two-part-chromosome2}). 
\citet{wang2020improved} improve the PGA by incorporating the reproduction mechanism into the invasive weed algorithm. 
They refer to their algorithm as RIPGA, which uses two-part chromosomes with breakpoints as an encoding method. 
In order to solve the M-mTSP, \citet{karabulut2021modeling} propose an Evolution Strategy (ES) with many-chromosome representation. 
The study utilizes a self-adaptive random local search with four different neighborhoods in order to enhance the solutions. 
The article by \citet{he2023memetic} proposes a Memetic Algorithm (MA) that can be utilized to solve the M-mTSP problem by using a generalized edge assembly crossover algorithm (EAX). 
In order to improve the solution, a variable neighborhood descent is utilized, and a thorough post-optimization is performed to further improve the solution. 
It should be noted that all algorithms that are designed for M-mTSP are capable of solving mTSP when the number of depots is set to one.

\begin{table}[h]
    \centering
    \caption{Summary of heuristic algorithms solving mTSP}
    
    \begin{tabular}{llll}
    \toprule
    \multicolumn{1}{l}{Algorithm}           & \multicolumn{1}{l}{Problem} & \multicolumn{1}{l}{Objective}    & \multicolumn{1}{l}{Approach} \\
    \midrule
    
    \citet{tang2000multiple}      & mTSP    & min-sum      & GA  \\
    \citet{park2001hybrid}    & VRP     & min-sum      & GA \\
    \citet{carter2006new}   & mTSP    & both         & GA \\
    \citet{junjie2006ant}   & mTSP    & min-sum      & AC  \\
    \citet{brown2007grouping}   & mTSP    & both         & GA   \\
    \citet{singh2009new}    & mTSP    & both         & GGA  \\
    \citet{liu2009ant}     & mTSP    & both         & AC \\
    \citet{chen2011operators}      & mTSP    & min-sum      & GA \\
    \citet{yuan2013new}      & mTSP    & both         & GA    \\
    \citet{li2013new}       & mTSP$^*$   & min-max      & GA  \\
    \citet{venkatesh2015two}  & mTSP    & both         & BC/IWA \\
    \citet{soylu2015general}    & mTSP    & both         & GVNS \\
    \citet{alves2015using}    & mTSP    & bi-objective & GA \\
    \citet{wang2017memetic}     & mTSP    & min-max      & MA  \\
    \citet{lu2019mission}      & mTSP    & min-max      & AC  \\
    \citet{jiang2020new}    & M-mTSP   & min-sum      & AC/PGA  \\
    \citet{wang2020improved}    & M-mTSP   & min-sum      & PGA/IWA \\
    \citet{karabulut2021modeling} & M-mTSP   & both         & ES \\
    \citet{liu2021comparative}      & VCMTSP  & min-sum      & GA  \\
    \citet{he2022hybrid}      & mTSP    & both         & TS/EAX  \\
    \citet{zheng2022effective}            & mTSP    & both         & VNS \\
    \citet{he2023memetic}   & M-mTSP & both & MA \\ 
    This study & mTSP & min-max & HGA/DP  \\
    \bottomrule
    \end{tabular}
    \label{tab:literature}
\end{table}

Lastly, we have summarized all the reviewed studies in Table \ref{tab:literature} for the convenience of the reader and listed them in chronological order.

\section{A Hybrid Genetic Algorithm} \label{sec:method}

This section outlines the structure of our algorithm, starting with the genetic algorithm.
The next step is to discuss the details of our dynamic programming-based \SPLIT{} algorithm and how it evaluates an individual.
We proceed to explore the various local search functions employed in the algorithm.
Lastly, we examine the implementation of removing intersections between tours and highlight its benefits to the min-max mTSP.

\begin{algorithm}
    \caption{Hybrid Genetic Algorithm} \label{alg:GA}
    \begin{algorithmic}[1]  
        \State $\Omega = \texttt{initial\_population()}$ \Comment{Section \ref{subsec:init}}  \label{GAline:init}
        \While{Stopping condition is not met}  \label{GAline:loopstart}
            \State $\texttt{sort}(\Omega)$   \Comment{Based on fitness and diversification factor}  \label{GAline:sort}
            \State Select $\omega_1$ and $\omega_2$ from $\Omega$   \label{GAline:parent}
            \State $c \gets \texttt{crossover}(\omega_1, \omega_2)$    \Comment{Section \ref{subsec:crossover}}   \label{GAline:crossover}
            {
            \State $\omega \gets \SPLIT{}(c)$  \Comment{Algorithms \ref{alg:split} and \ref{alg:extract}} \label{GAline:split}}
            \State $\texttt{educate}(\omega)$ 		\Comment{Section \ref{subsec:educate}}   \label{GAline:improvement}
            \State $\Omega_\textsc{f} \gets \Omega_\textsc{f} \cup \{\omega\}$    \label{GAline:add}
            \If{ $\texttt{size}(\Omega) = \mu+\lambda $}       \label{GAline:survivestart}
                \State $\texttt{select\_survivors}(\Omega)$
            \EndIf   \label{GAline:surviveend}
            \If{$\texttt{best}(\Omega)$ not improved for $It_{\textsc{DIV}}$ iterations}  \label{GAline:diversifystart}
                \State $\texttt{diversify}(\Omega)$
            \EndIf   \label{GAline:diversifyend}
        \EndWhile  \label{GAline:loopend}
        \State Return $\texttt{best}(\Omega)$
    \end{algorithmic}
\end{algorithm}

A simplified version of \citet{vidal2012hybrid}'s GA structure is used in this study. 
While our GA does not have any infeasible regions, the population control is similar to \citet{vidal2012hybrid}, in which the number of individuals varies between $\mu$ and $\mu + \lambda$. 
An overview of the procedure is presented in Algorithm \ref{alg:GA}.
The initial population is generated by producing $\mu$ individuals as described in Section \ref{subsec:init} (line \ref{GAline:init}).
The following steps are repeated until the stopping condition is met (lines \ref{GAline:loopstart}-\ref{GAline:loopend}).
The algorithm terminates when either the number of iterations without improvement reaches $It_{\textsc{ni}}$ or the total running time of the algorithm reaches the predetermined cutoff time (line \ref{GAline:loopstart}).

First, the population is sorted based on fitness and diversification factor (line \ref{GAline:sort}), which is explained in Section \ref{subsec:split}. 
Using the Tournament Selection method, two individuals are selected as parents from the population (line \ref{GAline:parent}). 
In order to select each parent, $k_{\textsc{tournament}}$ individuals are randomly selected, and the best of them is chosen. 
Upon selection of two parents from the population, a Similar Tour Crossover (STX) is performed on them in order to create the child's chromosome (line \ref{GAline:crossover}). 
More detailed information about STX crossover can be found in Section \ref{subsec:crossover}. 
The generated chromosome is a permutation of all cities that must be converted into a feasible mTSP solution. 
The \SPLIT{} algorithm is employed on the chromosome to divide the given sequence into $m$ tours in an optimal manner (line \ref{GAline:split}).
Next, the child undergoes a combination of local search functions and improvements before being added to the population (line \ref{GAline:improvement}). 
If the population size reaches $\mu + \lambda$, the individuals will be sorted in accordance with their fitness (min-max objective), the first $\mu$ individuals will survive, while the remainder will be discarded (lines \ref{GAline:survivestart}-\ref{GAline:surviveend}). 
In order to prevent the algorithm from becoming trapped in a local optimum, when there are no improvements after $It_{\textsc{div}}$ generations, an attempt is made to diversify the population by keeping $n_{\textsc{best}}$ number of individuals and adding new chromosomes similar to the generation of the initial population to increase the population size to $\mu$ again (lines \ref{GAline:diversifystart}-\ref{GAline:diversifyend}).

Following is a summary of the remainder of this section:
In Section \ref{subsec:split}, we introduce our dynamic programming algorithm, \SPLIT{}, which is specifically designed for individual evaluation.
Section \ref{subsec:init} describes the process of generating the initial population.
Section \ref{subsec:crossover} provides an explanation of the STX crossover.
Lastly, in Section \ref{subsec:educate}, we discuss the steps involved in educating the child.

\subsection{Individual evaluation by \SPLIT{}} \label{subsec:split}

\begin{figure}
    \resizebox{6in}{!}{%
        \begin{tikzpicture}
            \node[depotnode] (n0) at (0,0) {0};
            \node[customernode] (n1) at (-2,-1) {1};
            \node[customernode] (n2) at (-2.5,1) {2};
            \node[customernode] (n3) at (-1,3) {3};
            \node[customernode] (n4) at (1,3) {4};
            \node[customernode] (n5) at (2.5,1) {5};
            \node[customernode] (n6) at (2,-2) {6};
            \node[customernode] (n7) at (0.5,-3) {7};
            \draw[->, thick] (n0) -- (n1); 
            \draw[->, thick] (n1) -- (n2); 
            \draw[->, thick] (n2) -- (n3);
            \draw[->, thick] (n3) -- (n4); 
            \draw[->, thick] (n4) -- (n5); 
            \draw[->, thick] (n5) -- (n6); 
            \draw[->, thick] (n6) -- (n7); 
            \draw[->, thick] (n7) -- (n0); 
            \matrix[below = 1 em of n7, tour] (T1) 
        {
           1 \& 
           2 \& 
           3 \&
           4 \&
           5 \&
           6 \&
           7 \\
        };
        \matrix[below = 2 em of T1, tour] (T2) 
        {
           0 \&
           1 \& 
           2 \& 
           3 \&
           4 \&
           5 \&
           6 \&
           7 \&
           0 \\
        }; 
        \node at ([xshift=-38pt] T1-1-1.west) {Chromosome};
        \node at ([xshift=-30pt] T2-1-1.west) {TSP tour};
        \draw[->, thick] (T1) -- (T2) node[right, midway] {adding depot};

        \node[depotnode] (n0p) at (9,0) {0};
        \node[customernode] (n1p) at (7,-1) {1};
        \node[customernode] (n2p) at (6.5,1) {2};
        \node[customernode] (n3p) at (8,3) {3};
        \node[customernode] (n4p) at (10,3) {4};
        \node[customernode] (n5p) at (11.5,1) {5};
        \node[customernode] (n6p) at (11,-2) {6};
        \node[customernode] (n7p) at (9.5,-3) {7};

        \draw[->, thick] (n0p) -- (n1p); 
        \draw[->, thick] (n1p) -- (n2p); 
        \draw[->, thick] (n2p) -- (n3p);
        \draw[->, thick] (n3p) -- (n0p);
        \draw[->, thick] (n0p) -- (n4p); 
        \draw[->, thick] (n4p) -- (n5p); 
        \draw[->, thick] (n5p) -- (n0p);
        \draw[->, thick] (n0p) -- (n6p); 
        \draw[->, thick] (n6p) -- (n7p); 
        \draw[->, thick] (n7p) -- (n0p);
        \draw[->, ultra thick] (3.5,0) -- (5.5,0) node[above, midway] {\SPLIT{}};
        \matrix[below = 1 em of n7p, tour] (T1p) 
        {
           1 \& 
           2 \& 
           3 \&
           4 \&
           5 \&
           6 \&
           7 \\
        }; 
        \draw[->, ultra thick] ([yshift=10pt]T1p-1-3.north east) -- (T1p-1-3.north east);
        \draw[->, ultra thick] ([yshift=-10pt]T1p-1-3.south east) -- (T1p-1-3.south east);
        \draw[->, ultra thick] ([yshift=10pt]T1p-1-5.north east) -- (T1p-1-5.north east);
        \draw[->, ultra thick] ([yshift=-10pt]T1p-1-5.south east) -- (T1p-1-5.south east);
        \matrix[below = 2 em of T1p, tour] (T2p) 
        {
           0 \&
           1 \& 
           2 \& 
           3 \&
           0 \\
        }; 
        \matrix[below = 0.1 em of T2p-1-2.south east, tour] (T3p) 
        {
           0 \&
           4 \& 
           5 \& 
           0 \\
        }; 
        \matrix[below = 0.1 em of T3p-1-2.south east, tour] (T4p) 
        {
           0 \&
           6 \& 
           7 \& 
           0 \\
        }; 
        \draw[->, thick] (T1p) -- (T2p) ;
        \node at ([xshift=-45pt] T2p-1-1.west) {mTSP solution};
        \end{tikzpicture}%
        
    }

\caption{Illustration of \SPLIT{} algorithm.}
\label{fig:split}
\end{figure}
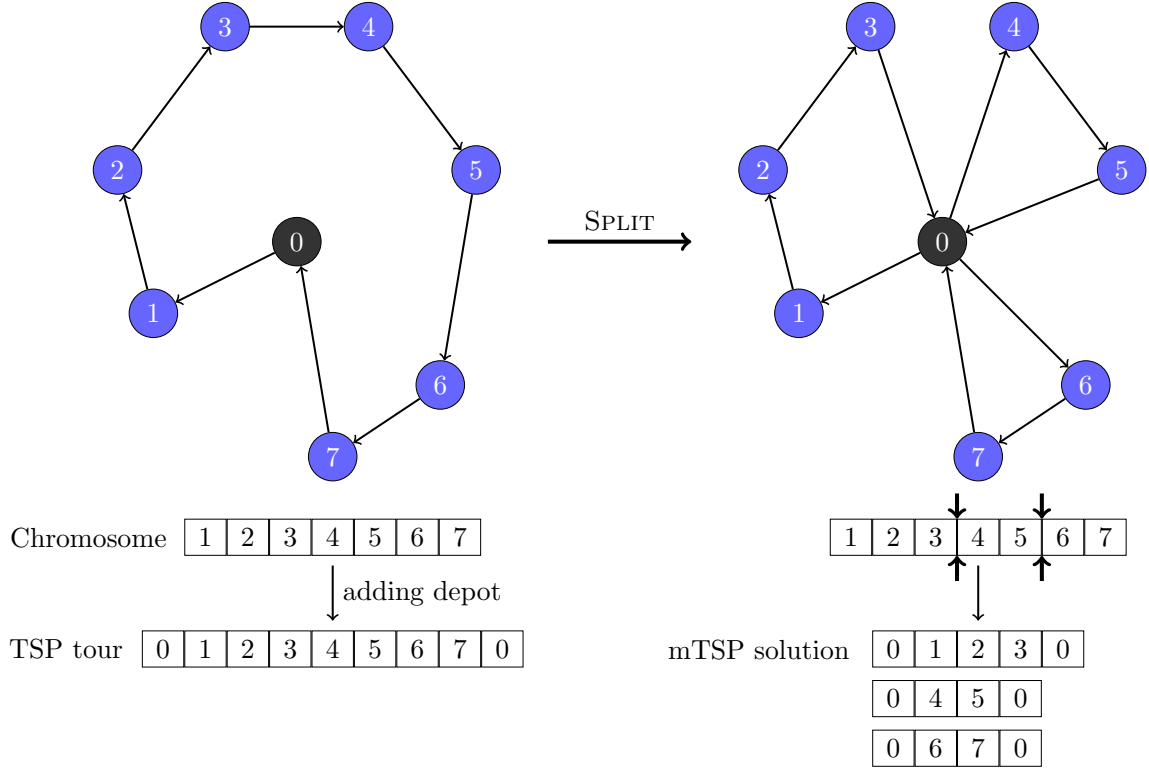

As previously mentioned, our GA utilizes a TSP sequence (excluding depot) as the chromosome representation. 
In this paper, we propose a dynamic programming algorithm that takes a TSP tour as input, identifies the delimiters, and then computes the optimal solution for the given sequence in the mTSP context.
As illustrated in Figure \ref{fig:split}, each chromosome representation is a TSP tour after adding the depot at the start and end. 
Without changing the order of the cities, the Split algorithm divides the initial sequence into m distinct sequences, each of which would become a tour after the depot is added. 

The \SPLIT{} algorithm, initially introduced by \citet{prins2004simple} for VRP to minimize total travel distance, has been adapted and modified for min-max mTSP in our study.
To facilitate understanding, we now provide the notation used in the algorithm:

\begin{itemize}
    \item $n$: the number of customer nodes
    \item $S=\{S_1, S_2, ... , S_n\}$: the given TSP sequence
    \item $m$: the number of salesmen
    \item $t_{S_i, S_j}$: travel time from node $S_i$ to node $S_j$
    \item $k$: position of the current node, $k \in \{1,2,...,n\}$
    \item $r$: the number of completed routes
    \item $R_k$: set of all possible numbers of completed routes at position $k$, where
\begin{equation*}
    R_k=
        \begin{cases}
             \{0\} & \text{ if } k=0,\\
             \{1,..., \min(k,m-1)\} & \text{ if } 1 \leq k < n, \\
             \{1, ..., m \} & \text{ if } k=n
        \end{cases}
\end{equation*}

    \item $V^k_r$: the makespan (min-max objective) of the path from the beginning depot to the node in position $k$, while completing $r$ routes
    \item $P^k_r$: the predecessor of $k$, while $r$ routes has been completed
\end{itemize}

\begin{algorithm}
    \caption{\SPLIT{} for \emph{min-max} mTSP} \label{alg:split}
    \begin{algorithmic}[1]
        \State $V^0_0 \gets 0 $
        \State $V^k_r \gets + \infty \quad \forall k=1$ to $n$ , $\forall r$ in $R_k$ 
        \For {$k=1$ to $n$}
            \For {$r \in R_{k-1}$}
                \If {$V^{k-1}_r < \infty$}
                    \State $T \gets 0; j \gets k$
                    \While{$j \leq n$} 
                        \If {k=j}
                            \State $T \gets t_{0,S_j} + t_{S_j,0}$   \Comment{$0$ represents depot.}
                        \Else 
                            \State $T \gets T - t_{S_{j-1},0} + t_{S_{j-1},S_j} + t_{S_j,0}$ 
                        \EndIf
                        \If {$r+1 \in R_j$}
                            \If {$\max(V^{k-1}_{r}, T) < V^j_{r+1}$}
                                \State $V^j_{r+1} \gets \max(V^{k-1}_{r}, T)$
                                \State $P^j_{r+1} \gets k-1$
                            \EndIf
                        \EndIf
                        \State $j \gets j+1$
                    \EndWhile
                \EndIf
            \EndFor
        \EndFor
    \end{algorithmic}
\end{algorithm}

\begin{algorithm}
    \caption{Extracting the mTSP solution} \label{alg:extract}
    \begin{algorithmic}[1]
        \For{$r=1$ to $m$}
            \State $tour(r) \gets \emptyset$
        \EndFor
        \State $r \gets m$
        \State $j \gets n$
        \While{$r>1$}
            \State $i \gets P^j_r$ 
            \For{$k=i+1:j$}
                \State $tour(r) \gets tour(r) \cup S_k$
            \EndFor
            \State $r \gets r-1$
            \State $j \gets i$
        \EndWhile
    \end{algorithmic}
\end{algorithm}

Algorithm \ref{alg:split} presents the details of the \SPLIT{} algorithm. 
Our objective is to minimize the longest tour from the depot to the $k$-th node in the given TSP sequence by constructing a series of $r$ tours.
For each step in the DP algorithm, we perform forward propagation on $V^k_r$ to generate a new tour and update the value of $V^j_{r+1}$ for all $j>k$. 
This process continues until all possible enumerations have been considered. 
The resulting $V^n_m$ yields the objective of optimal mTSP solution for the given sequence.
Additionally, we utilize $P^k_r$ for extracting the mTSP solution through Algorithm \ref{alg:extract}.

The optimal objective value of \SPLIT{} is the minimum time of the longest tour in the mTSP solution of an individual in the population. 
To promote diversity in the population, we calculate the fitness measure for each individual using the min-max objective function and the diversity factor.
According to \citet{vidal2012hybrid}, the normalized Hamming distance $\delta^H(P_1, P_2)$ can be calculated as follows:
\[\delta^H(P_1, P_2)=\frac{1}{n}\sum_{i=1}^{n}\boldsymbol{1}(P_1[i]\ne P_2[i]),\]
where $\boldsymbol{1}(\cdot)$ equals 1 if the condition specified within the parentheses is true and 0 otherwise, and $n$ represents the number of customer nodes.
The distance is measured between zero and one, where a value of one indicates that the representations of two individuals are completely different, and a value of zero indicates that the representations are similar.
To quantify diversity, we calculate the diversity contribution $\Delta(P)$ based on the average Hamming distance between an individual $P$ and its two closest neighbors in the population.
The fitness function of each individual is calculated using the following formula:
\[\text{fitness}(P) = \text{min-max}(P) \times \Big(1-\frac{n_{\textsc{Elite}}}{n_{\textsc{Population}}} \Big)^{\Delta(P)},\]
where we let $\text{min-max}(P)$ be the optimal minimized value of the longest tour found by the \SPLIT{} algorithm for individual $P$.

\subsection{Initial population} \label{subsec:init}

The \SPLIT{} algorithm can be employed on a TSP tour to generate an mTSP solution. 
It is expected that different TSP tours will result in different mTSP solutions, and, thus a more diverse population. 
This study utilizes four TSP algorithms: an exact algorithm implemented in \emph{Concorde} \citep{applegate2002solution} and three simple heuristic algorithms of \emph{nearest insertion, farthest insertion}, and \emph{cheapest insertion}. 
To create the remainder of the initial population, we randomly select one of the available TSP tours, modify it, and then apply the \SPLIT{} algorithm to obtain the individual. 
Changing the TSP tour is accomplished in one of the following ways: 

\begin{itemize}
    \item By selecting two randomly chosen positions on the tour, the nodes between these positions are inversed.
    \item Take the TSP tour and randomly divide it into $m$ subtours, and then shuffle the order of the subtours.
    \item Select a random number $r$ between $2$ and $\displaystyle\frac{n}{2}$, then randomly select $r$ positions and shuffle their order. 
\end{itemize}
Until $\mu$ individuals are present in the initial population, this process is repeated. 

\subsection{Similar tour crossover (STX)} \label{subsec:crossover}

\begin{figure}
    \centering
    \resizebox{6in}{!}{%
        \begin{tikzpicture}
            \matrix[chromosome] (P1) 
            {
                |[fill=white!80!cyan ]|   1 \& 
                |[fill=white!80!cyan ]|   2 \& 
                |[fill=white!80!cyan ]|   3 \&
                |[fill=white!80!cyan ]|   4 \&
                |[fill=white!80!cyan ]|   5 \&
                |[fill=white!80!cyan ]|   6 \&
                |[fill=white!80!cyan ]|   7 \&
                |[fill=white!80!cyan ]|   8 \&
                |[fill=white!80!cyan ]|   9 \&
                |[fill=white!80!cyan ]|   10 \&
                |[fill=white!80!cyan ]|   11 \&
                |[fill=white!80!cyan ]|   12 \\
            };
            \matrix[below = 1em of P1, chromosome] (P2) 
            {
                |[fill=white!80!gray]|   7 \& 
                |[fill=white!80!gray]|    4 \& 
                |[fill=white!80!gray]|    2 \&
                |[fill=white!80!gray]|    5 \&
                |[fill=white!80!gray]|    12 \&
                |[fill=white!80!gray]|    8 \&
                |[fill=white!80!gray]|    11 \&
                |[fill=white!80!gray]|    10 \&
                |[fill=white!80!gray]|    9 \&
                |[fill=white!80!gray]|    3 \&
                |[fill=white!80!gray]|    1 \&
                |[fill=white!80!gray]|    6 \\
            };
            \node at ([xshift=25pt] P1-1-12.east) {Parent 1};
            \node at ([xshift=25pt] P2-1-12.east) {Parent 2};
            \draw[ultra thick] ([yshift=10pt] P1-1-3.north east) -- ([yshift=-10pt] P1-1-3.south east);
            \draw[ultra thick] ([yshift=10pt] P1-1-8.north east) -- ([yshift=-10pt] P1-1-8.south east);
            \draw[ultra thick] ([yshift=10pt] P2-1-4.north east) -- ([yshift=-10pt] P2-1-4.south east);
            \draw[ultra thick] ([yshift=10pt] P2-1-9.north east) -- ([yshift=-10pt] P2-1-9.south east);

            \matrix[below = 2.5em of P2-1-3, chromosome] (T11) 
            {
                |[fill=white!80!cyan ]|    4 \&
                |[fill=white!80!cyan ]|    5 \&
                |[fill=white!80!cyan ]|    6 \&
                |[fill=white!80!cyan ]|    7 \&
                |[fill=white!80!cyan ]|    8 \\
            };
            \matrix[below = 0.3 em of T11-1-2.south east, chromosome] (T21) 
            {
                |[fill=white!80!gray]|   7 \& 
                |[fill=white!80!gray]|   4 \& 
                |[fill=white!80!gray]|   2 \&
                |[fill=white!80!gray]|   5 \\
            };
            \matrix[right = 3em of T11-1-4.east, chromosome] (T12) 
            {
                |[fill=white!80!cyan ]|    1 \&
                |[fill=white!80!cyan ]|    2 \&
                |[fill=white!80!cyan ]|    3 \\
            };
            \matrix[below = 0.3 em of T12-1-2.south, chromosome] (T22) 
            {
                |[fill=white!80!gray]|   3 \& 
                |[fill=white!80!gray]|   1 \& 
                |[fill=white!80!gray]|   6 \\
            };
            \matrix[right = 1em of T12-1-3.east, chromosome] (T13) 
            {
                |[fill=white!80!cyan ]|    9 \&
                |[fill=white!80!cyan ]|    10 \&
                |[fill=white!80!cyan ]|    11 \&
                |[fill=white!80!cyan ]|    12 \\
            };
            \matrix[below = 0.3 em of T13-1-3.south, chromosome] (T23) 
            {
                |[fill=white!80!gray]|   12 \& 
                |[fill=white!80!gray]|   8 \& 
                |[fill=white!80!gray]|   11 \&
                |[fill=white!80!gray]|   10 \&
                |[fill=white!80!gray]|   9 \\
            };
            \node at ([xshift=50pt] T13-1-4.south east) {Similar tours};
            \draw[dashed, thick, red] ([yshift=5pt] T11-1-2.north west) -- ([yshift=-5pt] T21-1-2.south west);
            \draw[dashed, thick, red] ([yshift=5pt] T11-1-3.north east) -- ([yshift=-5pt] T21-1-3.south east);
            \draw[dashed, thick, red] ([yshift=5pt] T12-1-2.north west) -- ([yshift=-5pt] T22-1-2.south west);
            \draw[dashed, thick, red] ([yshift=5pt] T12-1-2.north east) -- ([yshift=-5pt] T22-1-2.south east);
            \draw[dashed, thick, red] ([yshift=5pt] T13-1-2.north west) -- ([yshift=-5pt] T23-1-2.south west);
            \draw[dashed, thick, red] ([yshift=5pt] T13-1-3.north east) -- ([yshift=-5pt] T23-1-3.south east);

            \matrix[below = 9.5 em of P2-1-7.south, chromosome] (C1) 
            {
                |[fill=white!80!cyan]|   4 \& 
                |[fill=white!80!gray]|   4 \& 
                |[fill=white!80!gray]|   2 \&
                |[fill=white!80!cyan]|   7 \&
                |[fill=white!80!cyan]|   8 \&
                |[fill=white!80!cyan]|   1 \&
                |[fill=white!80!gray]|   1 \&
                |[fill=white!80!cyan]|   3 \&
                |[fill=white!80!gray]|   12 \&
                |[fill=white!80!cyan]|   10 \&
                |[fill=white!80!cyan]|   11 \&
                |[fill=white!80!gray]|   10 \&
                |[fill=white!80!gray]|   9 \\
            };
            \node at ([xshift=52pt] C1-1-13.east) {Child (unprocessed)};

            \matrix[below = 2.5 em of C1-1-5.south east, chromosome] (C2) 
            {
                |[fill=white!80!cyan]|   4 \& 
                |[fill=white!80!gray]|   2 \&
                |[fill=white!80!cyan]|   7 \&
                |[fill=white!80!cyan]|   8 \&
                |[fill=white!80!cyan]|   1 \&
                |[fill=white!80!cyan]|   3 \&
                |[fill=white!80!gray]|   12 \&
                |[fill=white!80!cyan]|   10 \&
                |[fill=white!80!cyan]|   11 \&
                |[fill=white!80!gray]|   9 \\
            };
            \node at ([xshift=55pt] C2-1-10.east) {repeated cities deleted};
            \matrix[below = 0.3 em of C2.south, chromosome] (R1) 
            {
                |[fill=white!80!red]|   5 \& 
                |[fill=white!80!red]|   6 \\
            };
            \node at ([xshift=42pt] R1-1-2.east) {Remaining cities};

            \matrix[below = 2.5 em of R1-1-2.south east, chromosome] (C3) 
            {
                |[fill=white!80!cyan]|   4 \& 
                |[fill=white!80!gray]|   2 \&
                |[fill=white!80!cyan]|   7 \&
                |[fill=white!80!red]|   5 \& 
                |[fill=white!80!cyan]|   8 \&
                |[fill=white!80!cyan]|   1 \&
                |[fill=white!80!cyan]|   3 \&
                |[fill=white!80!gray]|   12 \&
                |[fill=white!80!red]|   6 \&
                |[fill=white!80!cyan]|   10 \&
                |[fill=white!80!cyan]|   11 \&
                |[fill=white!80!gray]|   9 \\
            };
            \node at ([xshift=45pt] C3-1-12.east) {Child (processed)};
            \node[shape=circle, draw] at ([xshift=-20pt] P1-1-1.south west) {1}; 
            \node[shape=circle, draw] at ([xshift=-20pt] T11-1-1.south west) {2};
            \node[shape=circle, draw] at ([xshift=-20pt] C1-1-1.west) {3}; 
            \node[shape=circle, draw] at ([xshift=-20pt] C2-1-1.south west) {4}; 
            \node[shape=circle, draw] at ([xshift=-20pt] C3-1-1.west) {5}; 

        \end{tikzpicture}%
        
    }
       \caption{Illustration of STX crossover for a problem with size $n=12$ and $m=3$.}
       \label{fig:STX}
\end{figure}
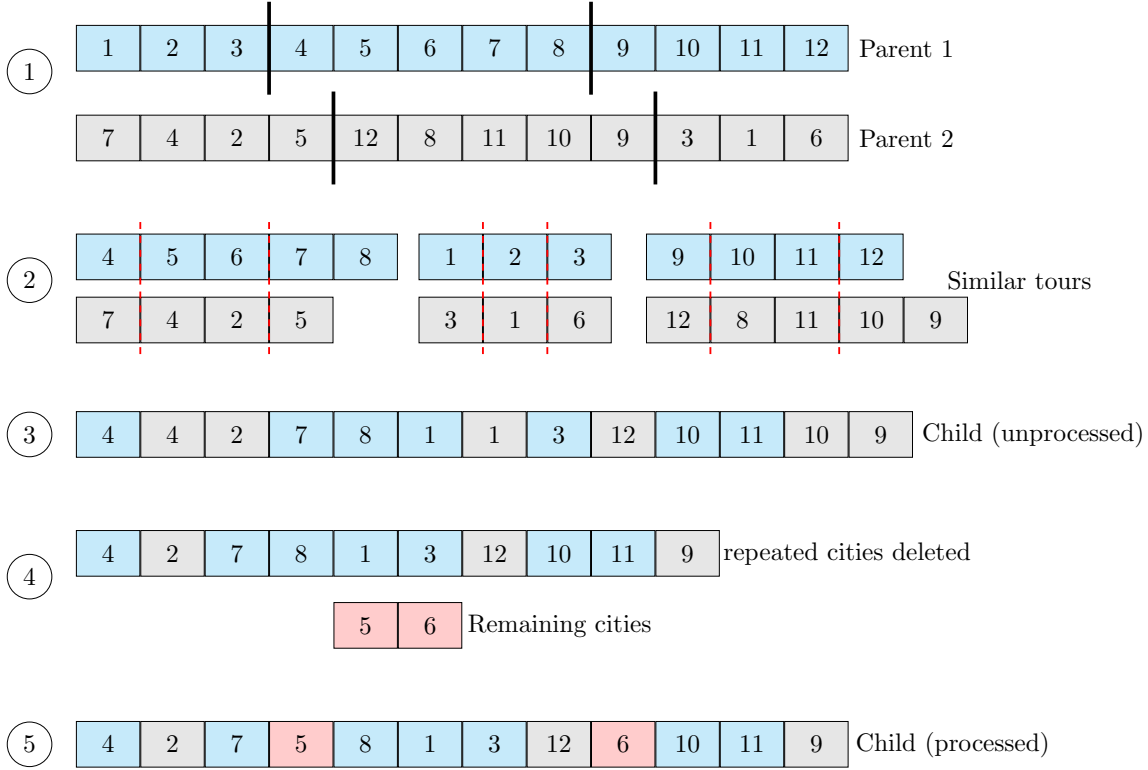

In order to replicate the mTSP solution from two parents, we develop a novel crossover function. 
Our proposed crossover works as follows: 
A tour is randomly chosen from the first parent, and then we select a tour from the second parent that has the maximum number of mutual cities with it. 
Afterward, a simple two-point crossover is implemented between selected tours, and the resulting tour is added to the child. 
In other words, two positions within the range of both tours are selected at random, and the resulting tour comprises the cities in between on the tour with a smaller number of cities along with the cities outside the chosen positions from the other tour. 
This process should be repeated until the child has $m$ tours. 
At the end of the process, some cities may appear more than once, and some may be absent from the solution. 
The repeating cities will be removed from the child's tour. 
In the final step, all remaining cities will be assigned to the child's tour based on a greedy approach. 
There should be an examination of all possible insertions for each city except for the longest tour, and the city should then be placed in the position with the least amount of increase. 
This crossover combines tours that share the most similarities in cities; hence the name Similar Tour Crossover (STX) is chosen. 
Figure \ref{fig:STX} illustrates a detailed example of the STX, which shows how it is employed to generate children from two parents. 
By conducting experiments on different instances, Section \ref{subsec:effect} examines the effectiveness of STX crossover.

\subsection{Chromosome education} \label{subsec:educate}
Our genetic algorithm is referred to as \emph{hybrid} because each offspring undergoes operations designed to improve its quality and, in general, to enhance the quality of the gene pool of the population. 
Three sequential layers are involved in the improvement of the offspring. 
At the first level, attention is paid to the geometry of the solution and the removal of intersections between different tours. 
The second layer attempts to make improvements on different tours regardless of the objective function, which is minimizing the longest tour. 
Lastly, the third layer focuses on reducing the length of the longest tour. 

\subsubsection*{Removing intersections}

\begin{figure}
    \centering
    \begin{subfigure}[b]{0.45\linewidth}
        \includegraphics[width=\linewidth]{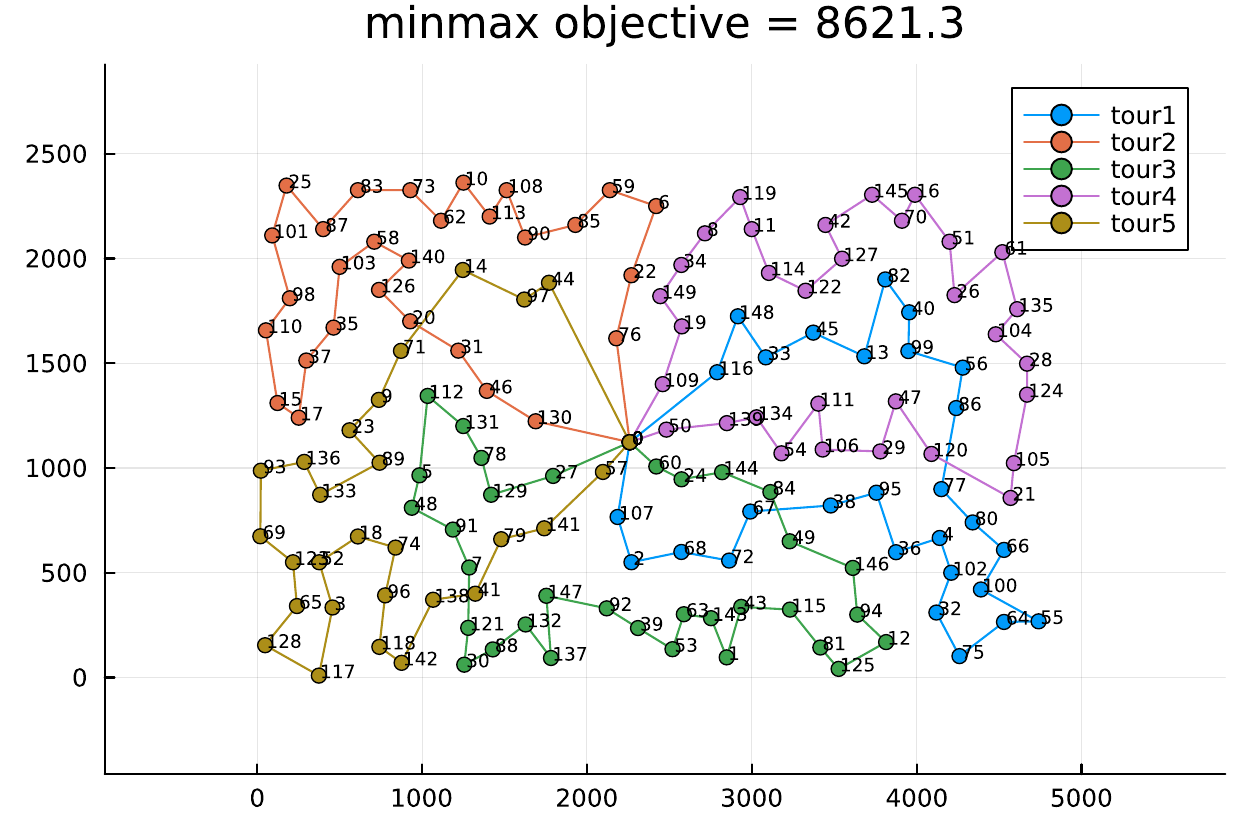}
        \caption{Before removing the intersections.}
        \label{fig:mtsp1}
    \end{subfigure}
    \begin{subfigure}[b]{0.45\linewidth}
        \includegraphics[width=\linewidth]{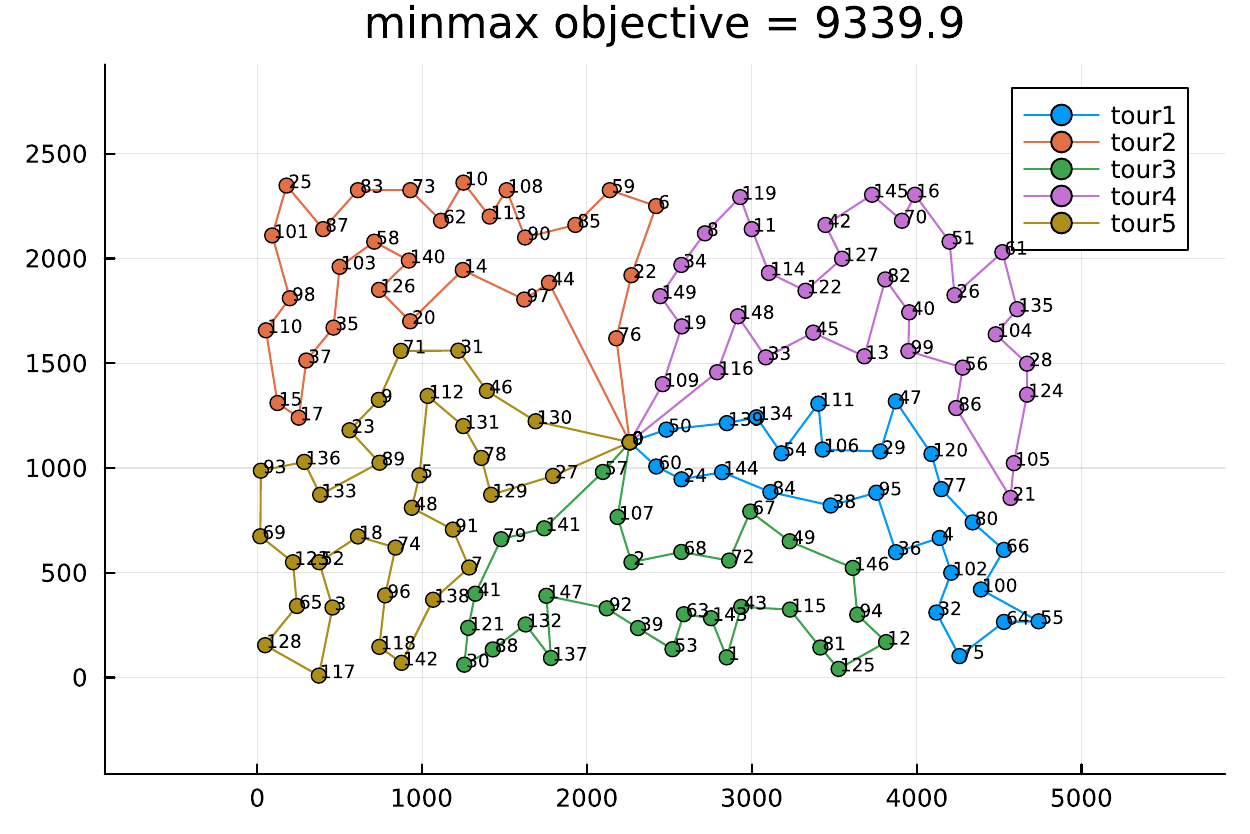}
        \caption{After removing the intersections.}
        \label{fig:mtsp2}
    \end{subfigure}
    \par\bigskip %
    \begin{subfigure}[b]{0.45\linewidth}
        \includegraphics[width=\linewidth]{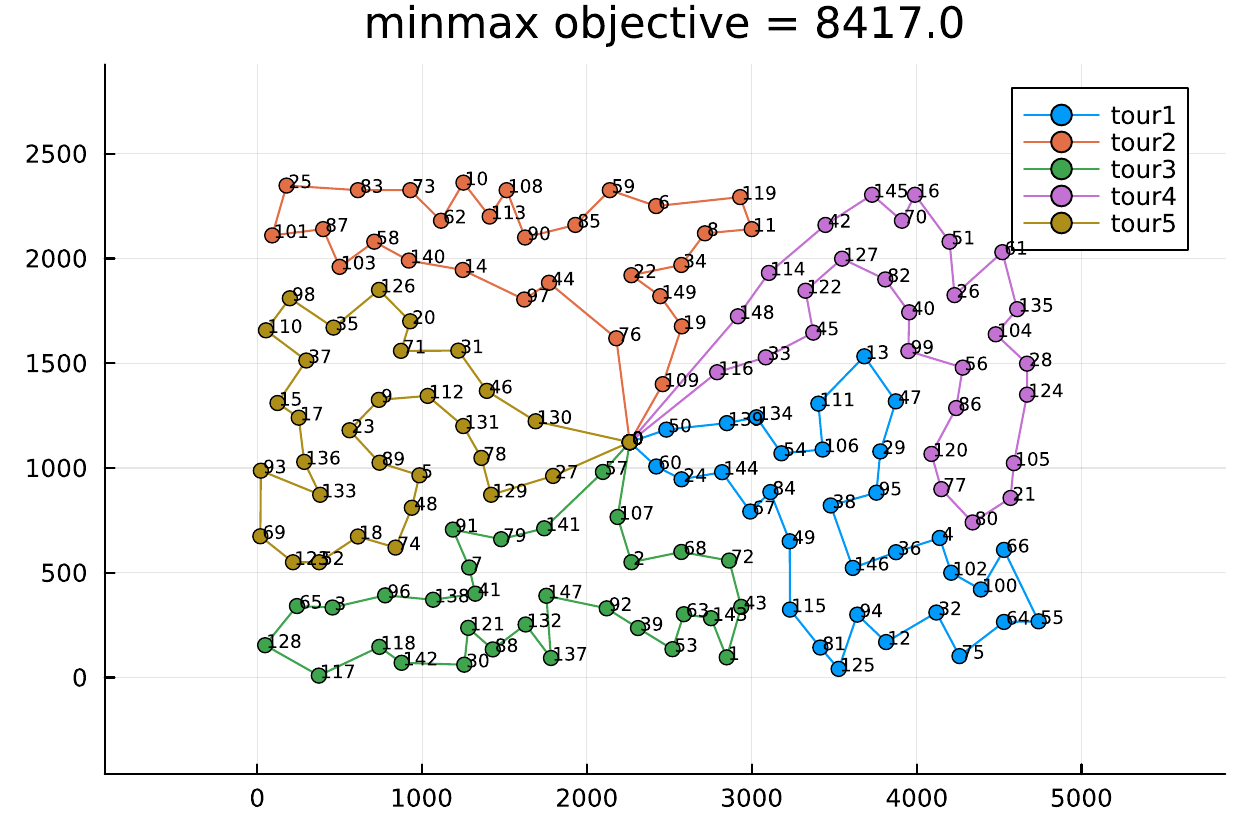}
        \caption{Best Known solution.}
        \label{fig:mtsp3}
    \end{subfigure}
    \caption{The importance of removing the intersections on instance MTSP-150 with $5$ salesmen.}
    \label{fig:intersection}
  \end{figure}

During this study, we discovered that for min-max mTSP, tours of optimal or near-optimal solutions tend to be separate and have the fewest intersections, particularly when the depot is in the center. 
Therefore, before beginning the local searches, with probability $P_{\textsc{remove}}$, a function is employed on the generated offspring that detects the intersections and removes them by exchanging sub-sequences of the intersected tours. 
An illustration of the implementation of this function can be observed in Figure \ref{fig:intersection}, which represents the intermediate and final solutions for the problem MTSP-150 with $m=5$ salesmen, from instance \emph{Set III} (Section \ref{subsec:set3}).
A diagram of the generated offspring in an intermediate generation, where the tours intersect, can be seen in Figure \ref{fig:mtsp1}. 
As an example, tour $1$ and tour $3$ intersect between cities $[38, 67]$ from tour $1$ and $[84, 49]$ from tour $3$. 
When this intersection is located, the function shifts the sub-sequence $[84, 144, 24, 60]$ from tour $3$ to tour $1$ as well as the sub-sequence $[67,72,68,2,107]$ from tour $1$ to tour $3$. 

The result of repeating this procedure for all intersections is an mTSP solution with no intersections, as shown in Figure \ref{fig:mtsp2}. 
It is important to note that it is very likely that the objective will worsen as a result of this function being employed. 
Nevertheless, implementing this technique once in a while may contribute to the gene pool moving toward the (near) optimal solution shown in Figure \ref{fig:mtsp3}. 
In addition, this function contributes to a more diverse solution, even for problems involving intersections in the optimal solution. 
Considering the drastic changes it makes to the solution, it can be considered a mutation. 
Getting out of the local optima can be accomplished by the sudden changes between tours that no local search can induce. 
A number of experiments are conducted in Section \ref{subsec:effect} in order to assess the contribution of removing the intersections.

\subsubsection*{Local searches}
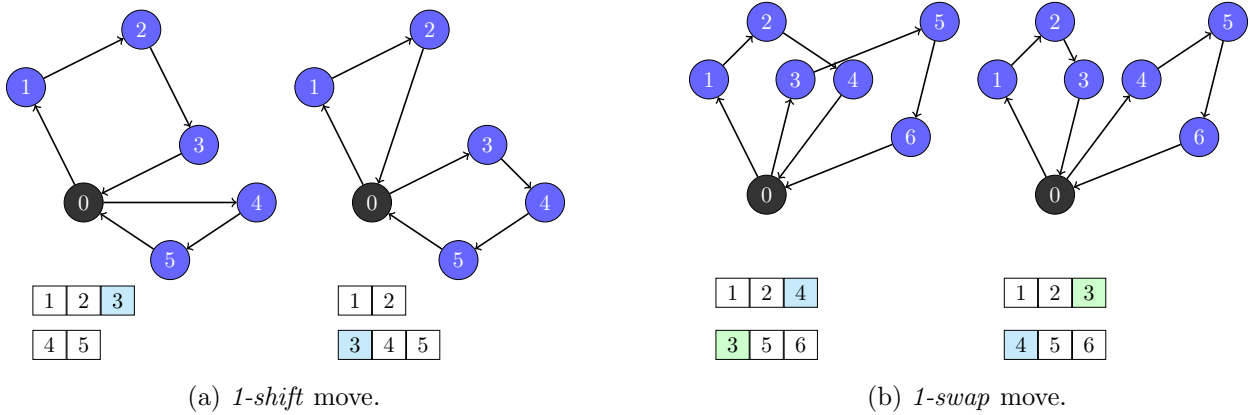
\begin{figure}
    \begin{subfigure}[b]{0.45\linewidth}\centering
        \resizebox{\linewidth}{!}{%
            \begin{tikzpicture}
                \node[depotnode] (n0) at (0,0) {0};
                \node[customernode] (n1) at (-1,2) {1};
                \node[customernode] (n2) at (1,3) {2};
                \node[customernode] (n3) at (2,1) {3};
                \node[customernode] (n4) at (3,0) {4};
                \node[customernode] (n5) at (1.5,-1) {5};
                \draw[->, thick] (n0) -- (n1); 
                \draw[->, thick] (n1) -- (n2); 
                \draw[->, thick] (n2) -- (n3);
                \draw[->, thick] (n3) -- (n0); 
                \draw[->, thick] (n0) -- (n4);
                \draw[->, thick] (n4) -- (n5);   
                \draw[->, thick] (n5) -- (n0); 
                \matrix[below = 2.5 em of n0, tour] (T1) 
            {
               1 \& 
               2 \& 
               |[fill=white!80!cyan]| 3 \\
            };
            \matrix[below = 0.3em of T1-1-1.south east, tour] (T2) 
            {
               4 \& 
               5 \\
            };

            \node[depotnode] (n0p) at (5,0) {0};
            \node[customernode] (n1p) at (4,2) {1};
            \node[customernode] (n2p) at (6,3) {2};
            \node[customernode] (n3p) at (7,1) {3};
            \node[customernode] (n4p) at (8,0) {4};
            \node[customernode] (n5p) at (6.5,-1) {5};
            \draw[->, thick] (n0p) -- (n1p); 
            \draw[->, thick] (n1p) -- (n2p); 
            \draw[->, thick] (n2p) -- (n0p);
            \draw[->, thick] (n0p) -- (n3p); 
            \draw[->, thick] (n3p) -- (n4p);
            \draw[->, thick] (n4p) -- (n5p);   
            \draw[->, thick] (n5p) -- (n0p); 
            \matrix[below = 2.5 em of n0p, tour] (T1p) 
            {
               1 \& 
               2 \\
            };
            \matrix[below = 0.3em of T1p-1-2.south, tour] (T2p) 
            {
                |[fill=white!80!cyan]| 3 \& 
               4 \&
               5 \\
            };
            \end{tikzpicture}%
            
        }
        \caption{\emph{1-shift} move.}
        \label{fig:1-shift}	
    \end{subfigure}
    \hfill
    \begin{subfigure}[b]{0.45\linewidth}\centering
        \resizebox{\linewidth}{!}{%
            \begin{tikzpicture}
                \node[depotnode] (n0) at (0,0) {0};
                \node[customernode] (n1) at (-1,2) {1};
                \node[customernode] (n2) at (0,3) {2};
                \node[customernode] (n3) at (0.5,2) {3};
                \node[customernode] (n4) at (1.5,2) {4};
                \node[customernode] (n5) at (3,3) {5};
                \node[customernode] (n6) at (2.5,1) {6};
                \draw[->, thick] (n0) -- (n1); 
                \draw[->, thick] (n1) -- (n2); 
                \draw[->, thick] (n2) -- (n4);
                \draw[->, thick] (n4) -- (n0); 
                \draw[->, thick] (n0) -- (n3);
                \draw[->, thick] (n3) -- (n5);   
                \draw[->, thick] (n5) -- (n6); 
                \draw[->, thick] (n6) -- (n0); 
                \matrix[below = 2.5 em of n0, tour] (T1) 
            {
               1 \& 
               2 \& 
               |[fill=white!80!cyan]| 4 \\
            };
            \matrix[below = 0.3em of T1.south, tour] (T2) 
            {
                |[fill=white!80!green]| 3 \& 
               5 \&
               6 \\
            };

            \node[depotnode] (n0p) at (5,0) {0};
            \node[customernode] (n1p) at (4,2) {1};
            \node[customernode] (n2p) at (5,3) {2};
            \node[customernode] (n3p) at (5.5,2) {3};
            \node[customernode] (n4p) at (6.5,2) {4};
            \node[customernode] (n5p) at (8,3) {5};
            \node[customernode] (n6p) at (7.5,1) {6};
            \draw[->, thick] (n0p) -- (n1p); 
            \draw[->, thick] (n1p) -- (n2p); 
            \draw[->, thick] (n2p) -- (n3p);
            \draw[->, thick] (n3p) -- (n0p); 
            \draw[->, thick] (n0p) -- (n4p);
            \draw[->, thick] (n4p) -- (n5p);   
            \draw[->, thick] (n5p) -- (n6p); 
            \draw[->, thick] (n6p) -- (n0p); 
            \matrix[below = 2.5 em of n0p, tour] (T1p) 
            {
               1 \& 
               2 \& 
               |[fill=white!80!green]| 3 \\
            };
            \matrix[below = 0.3em of T1p.south, tour] (T2p) 
            {
                |[fill=white!80!cyan]| 4 \& 
               5 \&
               6 \\
            };
            \end{tikzpicture}%
            
        }
        \caption{\emph{1-swap} move.}
        \label{fig:1-swap}	
    \end{subfigure}

    \caption{Inter-route neighborhoods.}
    \label{fig:neighborhood1}
\end{figure}

We now discuss how our algorithm works to perform local searches. 
The most commonly used neighborhoods in the mTSP (and VRP) can be divided into two categories, inter-tour neighborhoods and intra-tour neighborhoods. 
In the inter-tour neighborhoods, improvements are made by making changes between tours, while in the intra-tour neighborhoods, improvements are made within the tours. 
From inter-tour neighborhoods, we use 1-shift move and 1-swap move in this study. 
The 1-shift move involves taking a city from one tour and relocating it to another tour for possible improvement (Figure \ref{fig:1-shift}). 
In a 1-swap move, two cities from two different tours are exchanged (Figure \ref{fig:1-swap}).

Our study utilizes Reinsert, Exchange, Or-opt2, Or-opt3, and 2-opt intra-tour neighborhoods out of many available. 
There is only one tour involved in these neighborhood functions. 
In a Reinsert move, one city is removed and then placed in a new location (Figure \ref{fig:reinsert}). 
The Exchange move swaps two cities within a tour (Figure \ref{fig:exchange}). 
The Or-opt2 and Or-opt3 moves are similar to Reinsert, except they remove two and three adjacent cities, respectively, and place them in a different location (Figure \ref{fig:Or-opt2} and \ref{fig:Or-opt3}). 
The 2-opt move reverses a sub-sequence within the tour (Figure \ref{fig:2-opt}).

\begin{figure}
    \begin{subfigure}[b]{0.45\linewidth}\centering
        \resizebox{\linewidth}{!}{%
            \begin{tikzpicture}
                \node[depotnode] (n0) at (0,0) {0};
                \node[customernode] (n1) at (0.5,2) {1};
                \node[customernode] (n2) at (2,2) {2};
                \node[customernode] (n3) at (3,0.5) {3};
                \node[customernode] (n4) at (1,1) {4};
                \node[customernode] (n5) at (1,-1) {5};
                \draw[->, thick] (n0) -- (n1); 
                \draw[->, thick] (n1) -- (n2); 
                \draw[->, thick] (n2) -- (n3);
                \draw[->, thick] (n3) -- (n4); 
                \draw[->, thick] (n4) -- (n5); 
                \draw[->, thick] (n5) -- (n0); 
                \matrix[below = 1 em of n5, tour] (T1) 
            {
               1 \& 
               2 \& 
               3 \&
               |[fill=white!80!cyan]| 4 \&
               5 \\
            };

            \node[depotnode] (n0p) at (5,0) {0};
            \node[customernode] (n1p) at (5.5,2) {1};
            \node[customernode] (n2p) at (7,2) {2};
            \node[customernode] (n3p) at (8,0.5) {3};
            \node[customernode] (n4p) at (6,1) {4};
            \node[customernode] (n5p) at (6,-1) {5};
            \draw[->, thick] (n0p) -- (n1p); 
            \draw[->, thick] (n1p) -- (n4p); 
            \draw[->, thick] (n4p) -- (n2p);
            \draw[->, thick] (n2p) -- (n3p); 
            \draw[->, thick] (n3p) -- (n5p); 
            \draw[->, thick] (n5p) -- (n0p); 
            \matrix[below = 1 em of n5p, tour] (T1p) 
            {
               1 \& 
               |[fill=white!80!cyan]| 4 \&
               2 \& 
               3 \&
               5 \\
            };
            \end{tikzpicture}%
            
        }
        \caption{\emph{Reinsert} move.}
        \label{fig:reinsert}	
    \end{subfigure}
    \hfill
    \begin{subfigure}[b]{0.45\linewidth}\centering
        \resizebox{\linewidth}{!}{%
            \begin{tikzpicture}
                \node[depotnode] (n0) at (0,0) {0};
                \node[customernode] (n3) at (0.7,1.5) {3};
                \node[customernode] (n2) at (2,1.5) {2};
                \node[customernode] (n1) at (2.7,0) {1};
                \node[customernode] (n4) at (2,-1.5) {4};
                \node[customernode] (n5) at (0.7,-1.5) {5};
                \draw[->, thick] (n0) -- (n1); 
                \draw[->, thick] (n1) -- (n2); 
                \draw[->, thick] (n2) -- (n3);
                \draw[->, thick] (n3) -- (n4); 
                \draw[->, thick] (n4) -- (n5); 
                \draw[->, thick] (n5) -- (n0); 
                \matrix[below = 1 em of n5, tour] (T1) 
            {
                |[fill=white!80!cyan]|1 \& 
               2 \& 
               |[fill=white!80!cyan]|3 \&
                4 \&
               5 \\
            };

            \node[depotnode] (n0p) at (5,0) {0};
            \node[customernode] (n3p) at (5.7,1.5) {3};
            \node[customernode] (n2p) at (7,1.5) {2};
            \node[customernode] (n1p) at (7.7,0) {1};
            \node[customernode] (n4p) at (7,-1.5) {4};
            \node[customernode] (n5p) at (5.7,-1.5) {5};
            \draw[->, thick] (n0p) -- (n3p); 
            \draw[->, thick] (n3p) -- (n2p); 
            \draw[->, thick] (n2p) -- (n1p);
            \draw[->, thick] (n1p) -- (n4p); 
            \draw[->, thick] (n4p) -- (n5p); 
            \draw[->, thick] (n5p) -- (n0p); 
            \matrix[below = 1 em of n5p, tour] (T1p) 
            {
                |[fill=white!80!cyan]|3 \& 
               2 \& 
               |[fill=white!80!cyan]|1\&
                4 \&
               5 \\
            };
            \end{tikzpicture}%
            
        }
        \caption{\emph{Exchange} move.}
        \label{fig:exchange}	
    \end{subfigure}
    \par\bigskip %
    \begin{subfigure}[b]{0.45\linewidth}\centering
        \resizebox{\linewidth}{!}{%
        \begin{tikzpicture}
            \node[depotnode] (n0) at (0,0) {0};
            \node[customernode] (n3) at (0.7,1.5) {3};
            \node[customernode] (n4) at (2,1.5) {4};
            \node[customernode] (n1) at (2.7,0) {1};
            \node[customernode] (n2) at (2,-1.5) {2};
            \node[customernode] (n5) at (0.7,-1.5) {5};
            \draw[->, thick] (n0) -- (n1); 
            \draw[->, thick] (n1) -- (n2); 
            \draw[->, thick] (n2) -- (n3);
            \draw[->, thick] (n3) -- (n4); 
            \draw[->, thick] (n4) -- (n5); 
            \draw[->, thick] (n5) -- (n0); 
            \matrix[below = 1 em of n5, tour] (T1) 
        {
            |[fill=white!80!cyan]|1 \& 
            |[fill=white!80!cyan]|2 \& 
           3 \&
            4 \&
           5 \\
        };

        \node[depotnode] (n0p) at (5,0) {0};
        \node[customernode] (n3p) at (5.7,1.5) {3};
        \node[customernode] (n4p) at (7,1.5) {4};
        \node[customernode] (n1p) at (7.7,0) {1};
        \node[customernode] (n2p) at (7,-1.5) {2};
        \node[customernode] (n5p) at (5.7,-1.5) {5};
        \draw[->, thick] (n0p) -- (n3p); 
        \draw[->, thick] (n3p) -- (n4p); 
        \draw[->, thick] (n4p) -- (n1p);
        \draw[->, thick] (n1p) -- (n2p); 
        \draw[->, thick] (n2p) -- (n5p); 
        \draw[->, thick] (n5p) -- (n0p); 
        \matrix[below = 1 em of n5p, tour] (T1p) 
        {
            3 \& 
           4 \& 
           |[fill=white!80!cyan]|1\&
           |[fill=white!80!cyan]| 2 \&
           5 \\
        };
        \end{tikzpicture}%
            
        }
        \caption{\emph{Or-opt2} move.}
        \label{fig:Or-opt2}	
    \end{subfigure}
    \hfill
    \begin{subfigure}[b]{0.45\linewidth}\centering
        \resizebox{\linewidth}{!}{%
        \begin{tikzpicture}
            \node[depotnode] (n0) at (0,0) {0};
            \node[customernode] (n4) at (0.7,1.5) {4};
            \node[customernode] (n1) at (2,1.5) {1};
            \node[customernode] (n2) at (2.7,0) {2};
            \node[customernode] (n3) at (2,-1.5) {3};
            \node[customernode] (n5) at (0.7,-1.5) {5};
            \draw[->, thick] (n0) -- (n1); 
            \draw[->, thick] (n1) -- (n2); 
            \draw[->, thick] (n2) -- (n3);
            \draw[->, thick] (n3) -- (n4); 
            \draw[->, thick] (n4) -- (n5); 
            \draw[->, thick] (n5) -- (n0); 
            \matrix[below = 1 em of n5, tour] (T1) 
        {
            |[fill=white!80!cyan]|1 \& 
            |[fill=white!80!cyan]|2 \& 
            |[fill=white!80!cyan]|3 \&
            4 \&
           5 \\
        };

        \node[depotnode] (n0p) at (5,0) {0};
        \node[customernode] (n4p) at (5.7,1.5) {4};
        \node[customernode] (n1p) at (7,1.5) {1};
        \node[customernode] (n2p) at (7.7,0) {2};
        \node[customernode] (n3p) at (7,-1.5) {3};
        \node[customernode] (n5p) at (5.7,-1.5) {5};
        \draw[->, thick] (n0p) -- (n4p); 
        \draw[->, thick] (n4p) -- (n1p); 
        \draw[->, thick] (n1p) -- (n2p);
        \draw[->, thick] (n2p) -- (n3p); 
        \draw[->, thick] (n3p) -- (n5p); 
        \draw[->, thick] (n5p) -- (n0p); 
        \matrix[below = 1 em of n5p, tour] (T1p) 
        {
            4 \& 
            |[fill=white!80!cyan]|1 \& 
           |[fill=white!80!cyan]|2\&
           |[fill=white!80!cyan]| 3 \&
           5 \\
        };
        \end{tikzpicture}%
            
        }
        \caption{\emph{Or-opt3} move.}
        \label{fig:Or-opt3}	
    \end{subfigure}
	\par\bigskip %
    \begin{subfigure}[b]{0.45\linewidth}\centering
        \resizebox{\linewidth}{!}{%
            \begin{tikzpicture}
                \node[depotnode] (n0) at (0,0) {0};
                \node[customernode] (n1) at (0.7,1.5) {1};
                \node[customernode] (n4) at (2,1.5) {4};
                \node[customernode] (n3) at (2.7,0) {3};
                \node[customernode] (n2) at (2,-1.5) {2};
                \node[customernode] (n5) at (0.7,-1.5) {5};
                \draw[->, thick] (n0) -- (n1); 
                \draw[->, thick] (n1) -- (n2); 
                \draw[->, thick] (n2) -- (n3);
                \draw[->, thick] (n3) -- (n4); 
                \draw[->, thick] (n4) -- (n5); 
                \draw[->, thick] (n5) -- (n0); 
                \matrix[below = 1 em of n5, tour] (T1) 
            {
                1 \& 
                |[fill=white!80!cyan]| 2 \& 
               |[fill=white!80!cyan]| 3 \&
               |[fill=white!80!cyan]| 4 \&
               5 \\
            };

            \node[depotnode] (n0p) at (5,0) {0};
            \node[customernode] (n1p) at (5.7,1.5) {1};
            \node[customernode] (n4p) at (7,1.5) {4};
            \node[customernode] (n3p) at (7.7,0) {3};
            \node[customernode] (n2p) at (7,-1.5) {2};
            \node[customernode] (n5p) at (5.7,-1.5) {5};
            \draw[->, thick] (n0p) -- (n1p); 
            \draw[->, thick] (n1p) -- (n4p); 
            \draw[->, thick] (n4p) -- (n3p);
            \draw[->, thick] (n3p) -- (n2p); 
            \draw[->, thick] (n2p) -- (n5p); 
            \draw[->, thick] (n5p) -- (n0p); 
            \matrix[below = 1 em of n5p, tour] (T1p) 
            {
                1 \& 
                |[fill=white!80!cyan]| 4 \& 
               |[fill=white!80!cyan]| 3\&
               |[fill=white!80!cyan]| 2 \&
               5 \\
            };
            \end{tikzpicture}%
            
        }
        \caption{\emph{2-opt} move.}
        \label{fig:2-opt}	
    \end{subfigure}

    \caption{Intra-route neighborhoods.}
    \label{fig:neighborhood2}
\end{figure}

The critical point now is how and in what order these neighborhoods are utilized. 
The improvement process is conducted sequentially using two different functions. 
The first function, which is the second layer of improvement, involves taking each tour and finding its neighbors based on the distances between their cities. 
Each city from the chosen tour is investigated for a 1-shift move as well as a 1-swap move for all possible positions and cities from neighboring tours. 
When the 1-shift move is utilized, it decreases the length of the tour from which the city is removed and increases the length of the tour to which the chosen city is added.
When the amount of decrease is greater than the amount of increase, and the length of the increased tour does not exceed the maximum tour length, the 1-shift move is implemented. 
The 1-swap move is implemented if both tour lengths decrease. 
Until no further improvements are achieved, this process is repeated. 
Upon completion of the 1-shift and 1-swap moves, we employ the 2-opt move on all tours for all pairs of cities until no further gains can be achieved. 
We refer to the first function as Enriching the chromosome since it adds value to the genes of an individual. 

All that remains is to explain the third layer of offspring improvement.
As soon as the enrichment process is completed, the second function is activated for further improvement. 
The second function employs the Reinsert, Exchange, Or-opt2, and Or-opt3 moves in a self-adaptive manner. 
By using a roulette wheel, the number of improvements made by each of these moves is counted, and the probability of choosing a move depends on the outcome of the roulette wheel. 
{
    By dividing the number of improvements occurring in each move by the total number of improvements, the roulette wheel probabilities are calculated. 
    As an example, if the number of improvements by the search functions is $100, 200, 200$, and $500$, then the roulette wheel would have probabilities of $0.1, 0.2, 0.2$, and $0.5$. 
    As a starting point, the counts of improvements for all the moves are initialized to $100$ in order to create fair competition. 
    }
Choosing a tour from the solution is required in order to implement each of these moves. 
Since the longest tour plays a critical role in the achievement of the min-max objective, improving the longest tour is of greater importance than improving the others. 
Improvements in non-dominant tours can, however, also contribute to the enhancement of the gene pool. 
As a result, when implementing any of these intra-tour moves, we either choose the longest tour or a random tour. 
Using this strategy, we are certain that at least half of the time will be spent on improving the objective function.
The second function is called Improving the chromosome. 
Since we only examine one move randomly in each implementation, the improving function repeats the roulette wheel and local search process multiple times for each child. 
It should be noted, however, that there is a potential trade-off here. 
A strong implementation of local search at the beginning would prevent GA from diversifying the population and would result in premature convergence. 
In contrast, if local search is not employed sufficiently, local optima are less likely to be escaped. 
Thus, if the number of generations with no improvement is less than $n_{\textsc{imprv}}$, the improving process must be repeated $n_{\textsc{local}}^1$ times; otherwise $n_{\textsc{local}}^2$ times, where $n_{\textsc{local}}^1 < n_{\textsc{local}}^2$. 
As a final note, it is important to note that the process of selecting cities for potential relocations is not based solely on randomness. 
Our method only selects randomly from the nearest $n_{\textsc{close}}$ cities instead of all of them in order to save time.  

\section{Computational results} \label{sec:results}

In order to evaluate the effectiveness of our HGA, four benchmark sets of instances are used. 
Using Julia programming language, the algorithm is implemented on a Mac computer with 16 GB of RAM and an Apple M1 processor.
The parameters used in our HGA are $\mu=10$, $\lambda=20$, $k_{\textsc{tournament}}=2$, $It_{\textsc{div}}=1000$, $n_{\textsc{best}}=0.2\mu$, $n_{\textsc{elite}}=0.2n_{\textsc{population}}$, 
$P_{\textsc{remove}}=0.1$, $n_{\textsc{imprv}}=100$, $n_{\textsc{local}}^1=100$, $n_{\textsc{local}}^2=1000$, $n_{\textsc{close}}=0.1n$. 

The following benchmark instances are used for evaluating our HGA and comparing it with the best existing methods in the literature: 
\begin{itemize}
    \item \emph{Set I}: The number of cities in instances of this set is $50, 100$, and $200$. 
    There are three different numbers of salesmen for each size, resulting in nine total settings. 
    Cities are randomly generated nodes. 

    \item \emph{Set II}: A total of 16 instances are included in this set, as defined by \citet{necula2015tackling}. 
    There are four TSP instances, eil51, berlin52, eil76, and rat99, from TSPLIB that are solved with $m=2,3,5$ and $7$ numbers of salesmen. 

    \item \emph{Set III}: A widely studied set of instances defined by \citet{carter2006new}, which includes three Euclidean two-dimensional symmetric TSP instances with $51, 100$, and $150$ cities. 
    Solving each instance for different numbers of salesmen (from $m = 3, 5, 10, 20$ and $30$) gives us 12 instances in total. 

    \item \emph{Set IV}: A set of 24 instances defined by \citet{wang2017memetic} that consists of six TSP instances, ch150, kroA200, lin318, att532, rat783 and cb1173 from TSPLIB. 
    In this set, $m = 3, 5, 10$ and $20$ salesmen are used.  
\end{itemize}

\subsection{Comparative results for Set I} \label{subsec:set1}

\begin{table}[]
    \caption{Results of HGA for random instances (\emph{Set I}).}
    \label{tab:set1}
    \begin{adjustbox}{max width=\textwidth}
        \begin{tabular}{cllrrrlrrrlrrr}
            \toprule
                               & $m$ &  & \multicolumn{3}{c}{5}                                                                        &  & \multicolumn{3}{c}{7}                                                                        &  & \multicolumn{3}{c}{10}                                                                       \\
                               \cmidrule{1-2}   \cmidrule{4-6}  \cmidrule{8-10}  \cmidrule{12-14}
                            $N$ & method            &  & \multicolumn{1}{l}{cost} & \multicolumn{1}{l}{gap   (\%)} & \multicolumn{1}{l}{time   (sec)} &  & \multicolumn{1}{l}{cost} & \multicolumn{1}{l}{gap   (\%)} & \multicolumn{1}{l}{time   (sec)} &  & \multicolumn{1}{l}{cost} & \multicolumn{1}{l}{gap   (\%)} & \multicolumn{1}{l}{time   (sec)} \\
			\midrule
            \multirow{5}{*}{50} & LKH-3             &  & \textbf{2.00}            & 0.00                           & 187.46                           &  & 1.95                     & 0.00                           & 249.31                           &  & 1.91                     & 0.00                           & 170.20                           \\
                                & OR-tools          &  & 2.04                     & 2.00                           & 3.24                             &  & 1.96                     & 0.51                           & 3.75                             &  & 1.91                     & 0.00                           & 3.67                             \\
                                & ScheduleNet      &  & 2.17                     & 8.50                           & 1.60                             &  & 2.07                     & 6.15                           & 1.67                             &  & 1.98                     & 3.66                           & 1.90                             \\
                                & NCE-mTSP          &  & 2.02                     & 1.00                           & 2.48                             &  & 1.96                     & 0.51                           & 2.50                             &  & 1.91                     & 0.00                           & 3.44                             \\
                               & HGA-avg           &  & \textbf{2.00}             & 0.00                           & 2.75                             &  & \textbf{1.92}            & \textbf{-1.53}                 & 2.39                             &  & \textbf{1.82}            & \textbf{-4.71}                 & 2.31                             \\
                               \midrule
                              & $m$ &  & \multicolumn{3}{c}{5}                                                                     &  & \multicolumn{3}{c}{10}                                                                    &  & \multicolumn{3}{c}{15}                                                                    \\
                              \cmidrule{1-2}   \cmidrule{4-6}  \cmidrule{8-10}  \cmidrule{12-14}
                              & method            &  & cost                     & \multicolumn{1}{l}{gap   (\%)} & time   (sec)                     &  & cost                     & \multicolumn{1}{l}{gap   (\%)} & time   (sec)                     &  & cost                     & \multicolumn{1}{l}{gap   (\%)} & time   (sec)                     \\
			\midrule                              
            100                 & LKH-3             &  & \textbf{2.20}            & 0.00                           & 262.85                           &  & 1.97                     & 0.00                           & 474.78                           &  & 1.98                     & 0.00                           & 378.90                           \\
                                & OR-tools          &  & 2.41                     & 9.55                           & 35.47                            &  & 2.03                     & 3.05                           & 45.40                            &  & 2.03                     & 2.53                           & 48.86                            \\
                                & ScheduleNet      &  & 2.59                     & 17.73                          & 14.84                            &  & 2.13                     & 8.12                           & 16.22                            &  & 2.07                     & 4.55                           & 20.02                            \\
                                & NCE-mTSP          &  & 2.24                     & 1.82                           & 16.36                            &  & 1.97                     & 0.00                           & 13.00                            &  & 1.98                     & 0.00                           & 23.37                            \\
                                & HGA-avg           &  & 2.21                     & 0.45                           & 8.06                             &  & \textbf{1.95}            & \textbf{-1.02}                 & 6.51                             &  & \textbf{1.96}            & \textbf{-1.01}                 & 6.58                             \\
                              \midrule
                              & $m$ &  & \multicolumn{3}{c}{10}                                                                    &  & \multicolumn{3}{c}{15}                                                                    &  & \multicolumn{3}{c}{20}                                                                    \\
                              \cmidrule{1-2}   \cmidrule{4-6}  \cmidrule{8-10}  \cmidrule{12-14}
                                & method            &  & cost                     & \multicolumn{1}{l}{gap   (\%)} & time   (sec)                     &  & cost                     & \multicolumn{1}{l}{gap   (\%)} & time   (sec)                     &  & \multicolumn{1}{l}{cost} & \multicolumn{1}{l}{gap   (\%)} & time   (sec)                     \\
			\midrule
            200                 & LKH-3             &  & 2.04                     & 0.00                           & 1224.40                          &  & \textbf{2.00}            & 0.00                           & 1147.13                          &  & \textbf{1.97}            & 0.00                           & 908.14                           \\
                                & OR-tools          &  & 2.33                     & 14.22                          & 675.79                           &  & 2.33                     & 16.50                          & 604.31                           &  & 2.37                     & 20.30                          & 649.17                           \\
                                & ScheduleNet      &  & 2.45                     & 20.10                          & 193.41                           &  & 2.24                     & 12.00                          & 213.07                           &  & 2.17                     & 10.15                          & 225.50                           \\
                                & NCE-mTSP          &  & 2.06                     & 0.98                           & 84.96                            &  & \textbf{2.00}            & 0.00                           & 84.28                            &  & 2.02                     & 2.54                           & 108.91                           \\
                                & HGA-avg           &  & \textbf{2.02}            & \textbf{-0.98}                 & 27.99                            &  & \textbf{2.00}            & 0.00                           & 23.32                            &  & 1.98                     & 0.51                           & 26.66                          \\
                              \bottomrule
        \end{tabular}   
    \end{adjustbox}
\end{table}

This set of instances contains three sizes, $50, 100$, and $200$ cities (including the depot).
The problem is solved with $m = 5, 7$ and $10$ for $n = 50$. 
In the case of $n=100$ and $n=200$, the number of salesmen is $5,10,15$ and $10,15,20$, respectively. 
$100$ instances are generated randomly for each combination using a uniform distribution over $[0,1]\times[0,1]$. 
Table \ref{tab:set1} presents the average results of all $100$ instances. 
The HGA is stopped when there are no improvements observed after $2500$ generations ($It_{\textsc{NI}}=2500$).
HGA has been implemented $10$ times for each instance; therefore, HGA-avg represents the average performance of HGA. 
LKH-3 \citep{helsgaun2017extension}, OR-tools \citep{perronor}, ScheduleNet \citep{parklearn} and NCE \citep{kim2022neuro} are used as baseline algorithms for comparison. 
HGA has obtained the best results in five problem sizes out of nine and equal results in two. 
Only in two problem sizes does HGA fail to defeat LKH-3. 
The difference in computational time, however, is extremely significant.

\subsection{Comparative results for Set II} \label{subsec:set2}

\begin{table}[]
    \begin{adjustbox}{max width=\textwidth}
        \begin{tabular}{l|rrrr|rrrr|rrrr|rrrr}
            \toprule
            Instance       & \multicolumn{4}{c|}{eil51}           & \multicolumn{4}{c|}{berlin52}         & \multicolumn{4}{c|}{eil76}            & \multicolumn{4}{c}{rat99}          \\
            \midrule
            m              & 2                & 3                & 5                & 7                & 2                & 3                & 5                & 7               & 2                & 3                & 5                & 7               & 2               & 3                & 5       & 7       \\
            \midrule
            CPLEX          & \textbf{222.7}   & \textbf{159.6}   & 124.0            & \textbf{112.1}   & \textbf{4110.2}  & 3244.4           & 2441.4           & \textbf{2440.9} & \textbf{280.9}   & 197.3            & 150.3            & 139.6           & 728.8           & 587.2            & 469.3            & 443.9   \\
            LKH3           & \textbf{222.7}   & \textbf{159.6}   & 124.0            & \textbf{112.1}   & \textbf{4110.2}  & 3244.4           & 2441.4           & \textbf{2440.9} & \textbf{280.9}   & 197.3            & 150.3            & 139.6           & 728.8           & 587.2            & 469.3            & 443.9   \\
            OR-Tools       & 243.3            & 170.5            & 127.5            & \textbf{112.1}   & 4665.5           & 3311.3           & 2482.6           & \textbf{2440.9} & 318.0            & 212.4            & 143.4            & 128.3           & 762.2           & 552.1            & 473.7            & 442.5   \\
            \midrule
            NCE            & 235.0            & 170.3            & 121.6            & \textbf{112.1}   & \textbf{4110.0}    & 3274.0             & 2660.0       & \textbf{2441.0} & 285.5            & 211.0            & 144.6            & \textbf{127.6}  & 695.8           & 527.8            & 458.6            & 441.6   \\
            NCE-mTSP       & 226.1            & 166.3            & 119.9            & \textbf{112.1}   & 4128.0             & 3191.0             & 2474.0       & \textbf{2441.0} & 282.1            & 197.5            & 147.2            & \textbf{127.6}  & \textbf{666.0}  & 533.2            & 462.2            & 443.9   \\
            \midrule
            BKS            & 222.7            & 159.6            & 119.9            & 112.1            & 4110.2           & 3191.0             & 2441.4         & 2440.9          & 280.9            & 197.3            & 143.4            & 127.6           & 666.0           & 527.8            & 458.6            & 441.6   \\
            \midrule
            HGA-best       & \textbf{222.7}   & \textbf{159.6}   & \textbf{118.1}   & \textbf{112.1}   & \textbf{4110.2}  & \textbf{3069.6}  & \textbf{2440.9}  & \textbf{2440.9} & \textbf{280.9}   & \textbf{196.7}   & \textbf{142.9}   & \textbf{127.6}  & \textbf{666.0}    & \textbf{517.7}   & \textbf{450.3}   & \textbf{436.7}   \\
            HGA-avg        & 222.9            & \textbf{159.6}   & \textbf{118.3}   & \textbf{112.1}   & \textbf{4110.2}  & \textbf{3073.8}  & \textbf{2440.9}  & \textbf{2440.9} & 281.6            & \textbf{196.7}   & 143.5            & \textbf{127.6}  & 668.0           & \textbf{517.8}   & \textbf{450.3}   & \textbf{437.3}   \\
            time (sec/run) & 4.0              & 3.2              & 3.7              & 2.7              & 4.1              & 3.3              & 2.9              & 2.8             & 5.2              & 5.0              & 6.4              & 4.7             & 8.0             & 6.7              & 9.7              & 16.2    \\
            \midrule
            best gap       & 0.00\%           & 0.00\%           & -1.50\%          & 0.00\%           & 0.00\%           & -3.80\%          & -0.02\%          & 0.00\%          & 0.00\%           & -0.30\%          & -0.35\%          & 0.00\%          & 0.00\%          & -1.91\%          & -1.81\%          & -1.11\% \\
            avg gap        & -1.42\%          & -4.03\%          & -1.33\%          & 0.00\%           & -0.43\%          & -3.67\%          & -1.34\%          & 0.00\%          & -0.18\%          & -0.41\%          & -2.51\%          & 0.00\%          & 0.30\%          & -2.89\%          & -2.57\%          & -1.49\% \\ 
            \bottomrule
            \end{tabular}
\end{adjustbox}
\caption{Results of HGA for \emph{Set II}. The `best gap' is the gap between HGA-best and BKS. The `average gap' is the gap between HGA-avg and NCE-mTSP.}
\label{tab:set2}
\end{table}

In this set of instances, four TSPLIB-based instances are solved with different numbers of salesmen: $m=2,3,5$, and $7$. 
The stopping condition for HGA is the same as Section \ref{subsec:set1}.
Ten runs of the HGA have been conducted for each instance, and the average and best results are reported. 
The baseline algorithms used for comparison include CPLEX, LKH-3, OR-tools, NCE, and NCE-mTSP \citep{kim2022neuro}.
Table \ref{tab:set2} presents the results of the HGA and baseline algorithms. 
The CPLEX results represent the best bounds determined by CPLEX. 
The results for the baseline algorithms are reported based on \citet{parklearn} and \citet{kim2022neuro}.
Since the running times of the baseline algorithms were not provided, only the running time of the HGA is reported.
Among the $16$ problems in this set, the HGA's solution is equal to or better than the baseline algorithms in all cases and strictly better in $8$ cases. 
To the best of our knowledge, the reported BKS (Best Known Solution) row in Table \ref{tab:set2} represents the best-known solutions to these problems. 
Therefore, the HGA has improved the BKS for exactly half of the problems in this set.

\subsection{Comparative results for Set III} \label{subsec:set3}

\begin{table}
    \centering
    \begin{adjustbox}{max width=\textwidth}
    \begin{tabular}{llrrrrrrrrrrrrrrr}
    \toprule
    & & \multicolumn{8}{c}{{integer distances}} & \multicolumn{7}{c}{{floating point distances}} \\
    \cmidrule(lr){3-10} \cmidrule(lr){11-17}
    \multicolumn{2}{c}{{Algorithm}} & & \multicolumn{1}{c}{{ES}} & \multicolumn{2}{c}{{ITSHA}} & \multicolumn{2}{c}{{HGA}} & \multicolumn{2}{c}{{Gap over}} & & \multicolumn{2}{c}{{HSNR}} & \multicolumn{2}{c}{{HGA}} & \multicolumn{2}{c}{{Gap over}} \\
    \cmidrule(lr){4-4} \cmidrule(lr){5-6} \cmidrule(lr){7-8} \cmidrule(lr){9-10} \cmidrule(lr){12-13} \cmidrule(lr){14-15} \cmidrule(lr){16-17}
    \multicolumn{1}{c}{{Instance}} & \multicolumn{1}{c}{{$m$}} & \multicolumn{1}{c}{{BKS}} & \multicolumn{1}{c}{{best}} & \multicolumn{1}{c}{{best}} & \multicolumn{1}{c}{{avg}} & \multicolumn{1}{c}{{best}} & \multicolumn{1}{c}{{avg}} & \multicolumn{1}{c}{{best}} & \multicolumn{1}{c}{{avg}} & \multicolumn{1}{c}{{BKS}} & \multicolumn{1}{c}{{best}} & \multicolumn{1}{c}{{avg}} & \multicolumn{1}{c}{{best}} & \multicolumn{1}{c}{{avg}} & \multicolumn{1}{c}{{best}} & \multicolumn{1}{c}{{avg}} \\
    \midrule
    {MTSP-51} & {3} & {159} & {160} & {\textbf{159}} & {159.0} & {\textbf{159}} & {159.0} & {0.00\%} & {0.00\%} & {159.57} & {\textbf{159.57}} & {159.99} & {\textbf{159.57}} & {159.57} & {0.00\%} & {-0.26\%} \\
    & {5} & {118} & {\textbf{118}} & {\textbf{118}} & {118.0} & {\textbf{118}} & {118.0} & {0.00\%} & {0.00\%} & {118.13} & {\textbf{118.13}} & {118.13} & {\textbf{118.13}} & {118.13} & {0.00\%} & {0.00\%} \\
    & {10} & {112} & {\textbf{112}} & {\textbf{112}} & {112.0} & {\textbf{112}} & {112.0} & {0.00\%} & {0.00\%} & {112.07} & {\textbf{112.07}} & {112.07} & {\textbf{112.07}} & {112.07} & {0.00\%} & {0.00\%} \\
    {MTSP-100} & {3} & {8507} & {8509} & {\textbf{8507}} & {8507.0} & {\textbf{8507}} & {8507.0} & {0.00\%} & {0.00\%} & {8509.16} & {\textbf{8509.16}} & {8578.51} & {\textbf{8509.16}} & {8509.16} & {0.00\%} & {-0.81\%} \\
    & {5} & {6766} & {\textbf{6766}} & {6770} & {6774.1} & {6769} & {6770.6} & {-0.01\%} & {-0.05\%} & {6765.73} & {\textbf{6765.73}} & {6774.24} & {6767.82} & {6770.50} & {0.03\%} & {-0.06\%} \\
    & {10} & {6358} & {\textbf{6358}} & {\textbf{6358}} & {6358.0} & {\textbf{6358}} & {6358.0} & {0.00\%} & {0.00\%} & {6358.49} & {\textbf{6358.49}} & {6358.49} & {\textbf{6358.49}} & {6358.49} & {0.00\%} & {0.00\%} \\
    & {20} & {6358} & {\textbf{6358}} & {\textbf{6358}} & {6358.0} & {\textbf{6358}} & {6358.0} & {0.00\%} & {0.00\%} & {6358.49} & {\textbf{6358.49}} & {6358.49} & {\textbf{6358.49}} & {6358.49} & {0.00\%} & {0.00\%} \\
    {MTSP-150} & {3} & {13039} & {13151} & {13084} & {13236.5} & \cellcolor{gray!50} {\textbf{13039}} & {13087.7} & {-0.34\%} & {-1.12\%} & {13038.30} & {13174.30} & {13352.37} & {13038.34} & {13093.48} & {-1.03\%} & {-1.94\%} \\
    & {5} & {8416} & {8466} & {8465} & {8543.6} & \cellcolor{gray!50} {\textbf{8416}} & {8500.5} & {-0.58\%} & {-0.50\%} & {8417.02} & {8479.60} & {8602.15} & {\textbf{8417.02}} & {8486.81} & {-0.74\%} & {-1.34\%} \\
    & {10} & {5557} & {\textbf{5557}} & {\textbf{5557}} & {5604.0} & {5564} & {5582.9} & {0.13\%} & {-0.38\%} & {5557.41} & {5616.71} & {5668.59} & {5561.97} & {5587.37} & {-0.97\%} & {-1.43\%} \\
    & {20} & {5246} & {\textbf{5246}} & {\textbf{5246}} & {5246.0} & {\textbf{5246}} & {5246.0} & {0.00\%} & {0.00\%} & {5246.49} & {\textbf{5246.49}} & {5246.49} & {\textbf{5246.49}} & {5246.49} & {0.00\%} & {0.00\%} \\
    & {30} & {5246} & {\textbf{5246}} & {\textbf{5246}} & {5246.0} & {\textbf{5246}} & {5246.0} & {0.00\%} & {0.00\%} & {5246.49} & {\textbf{5246.49}} & {5246.49} & {\textbf{5246.49}} & {5246.49} & {0.00\%} & {0.00\%} \\
    \midrule
    {Average} & & {5490} & {5504} & {5498} & {5521.9} & {5491} & {5503.8} & {-0.07\%} & {-0.17\%} & {5490.61} & {5512.10} & {5548.00} & {5491.17} & {5503.92} & {-0.23\%} & {-0.49\%} \\
    \bottomrule
    \end{tabular}
    \end{adjustbox}
    \caption{{Results of HGA for \emph{Set III}.
    Bold values in the `best' columns are the values equal to BKS, which are the best-known solutions in the literature regardless of the computational time.
    The shaded values are the new best solutions achieved by HGA.}}
    \label{tab:set3}
\end{table}

This set of instances, proposed by \citet{carter2006new}, consists of three instances named MTSP-51, MTSP-100, and MTSP-150. 
MTSP-51 is solved with $m=3,5$ and $10$ salesmen, MTSP-100 is solved with $m=3,5,10$ and $20$, and MTSP-150 is solved with $m=3,5,10,20$ and $30$ salesmen.
Table \ref{tab:set3} presents detailed results for the HGA and its comparison with baseline algorithms. 
The following algorithms are used as baselines: ES \citep{karabulut2021modeling}, HSNR \citep{he2022hybrid}, and ITSHA \citep{zheng2022effective}. 
ES and ITSHA are run $10$ times with a cutoff time of $n$ seconds to stop the algorithm, where $n$ is the number of cities. 
HSNR is tested on this set's instances $20$ times with two different cutoff times: $(1.36 \times n)/5$ seconds and $(n/100)\times 4$ minutes ($2.4 \times n$ seconds). 
To provide a fair comparison, the results are compared with the $(1.36 \times n)/5$ seconds cutoff time of HSNR.
In this study, the HGA is implemented on each instance $10$ times with a cutoff time of $n/5$ seconds. 
The decision to use a cutoff time of $n/5$ seconds for the HGA was based on the aim of demonstrating its fast convergence capability.

{
With the exception of \citet{he2022hybrid} and \citet{he2023memetic}, all the studies utilize integer distances when analyzing this set of instances.
\citet{he2022hybrid} and \citet{he2023memetic} used floating point distances. 
Therefore, we decided to solve the instances of this set with both integer and floating point distances. 
For ES, only the best results are displayed in Table \ref{tab:set3}. 
As for ITSHA and HSNR, both the best and average results are reported, and HGA's gaps are compared to these results. 
According to the table, HGA outperforms the baseline algorithms in terms of best and average solutions. 
In $10$ of $12$ instances, HGA achieves results equal to BKS when the distance is an integer. 
Additionally, it improves the BKS in two instances. 
When compared with ITSHA's average results, HGA's average results are equal or better in all instances.
The HGA achieves results that are equal to the BKS in $9$ out of $12$ instances when considering floating point distances.
The average results of HGA are equivalent or superior to those of HSNR in all instances.
It is worthy of note that \citet{he2023memetic} contributed to finding some of the best-known solutions in the second BKS column. 
However, since they implemented MA with a cutoff time of $2.4 \times n$ seconds, their results were not included in the table. 
Nevertheless, their best result for MTSP-150 is identical to ours for $m=3$ and almost the same as ours for $m=5$. 
In the latter case, MA obtained $13038.30$ and HGA obtained $13038.34$. 
We can still claim that our results represent the new best solutions with a cutoff time of $n/5$ seconds, as HGA achieved these solutions $12$ times faster.
}

\subsection{Comparative results for Set IV} \label{subsec:set4}

\begin{table}[]
    \begin{adjustbox}{max width=\textwidth}
        \begin{tabular}{lrrrrrrrrrrr}
            \toprule
            \multicolumn{3}{c}{Algorithm}                                                  & \multicolumn{1}{c}{MASVND} & \multicolumn{1}{c}{ES}    & \multicolumn{1}{c}{HSNR}      & \multicolumn{2}{c}{ITSHA}                       & \multicolumn{2}{c}{HGA}               & \multicolumn{2}{c}{Gap over}                       \\
            \midrule
            \multicolumn{3}{c}{cutoff time (seconds)}                                      & \multicolumn{1}{c}{$n/5$}  & \multicolumn{1}{c}{$n/5$} & \multicolumn{1}{c}{$1.36n/5$} & \multicolumn{2}{c}{$n/5$}                   & \multicolumn{2}{c}{$n/5$}                    & \multicolumn{2}{c}{ITSHA}                          \\
            \cmidrule(lr){4-4}   \cmidrule(lr){5-5}  \cmidrule(lr){6-6}  \cmidrule(lr){7-8} \cmidrule(lr){9-10}  \cmidrule(lr){11-12}   
            \multicolumn{1}{c}{Instance} & \multicolumn{1}{c}{$m$} & \multicolumn{1}{c}{BKS} & \multicolumn{1}{c}{best}  & \multicolumn{1}{c}{best}  & \multicolumn{1}{c}{best}     & \multicolumn{1}{c}{best} & \multicolumn{1}{c}{avg} & \multicolumn{1}{c}{best} & \multicolumn{1}{c}{avg} & \multicolumn{1}{c}{Best} & \multicolumn{1}{c}{avg} \\
            \midrule
            ch150                        & 3                     & 2401.63                 & 2429.49                    & 2407.59                   & 2425.87                       & 2405.94                  & 2435.25                 & \textbf{2401.63}         & 2405.79                 & -0.18\%                  & -1.21\%                 \\
                                         & 5                     & 1740.63                 & 1758.08                    & 1741.61                   & 1744.26                       & \textbf{1740.63}         & 1765.62                 & 1741.13                  & 1741.82                 & 0.03\%                   & -1.35\%                 \\
                                         & 10                    & 1554.64                 & \textbf{1554.64}           & \textbf{1554.64}          & \textbf{1554.64}              & \textbf{1554.33}         & 1554.33           & \textbf{1554.64}         & 1554.64                 & 0.00\%                   & 0.00\%                  \\
                                         & 20                    & 1554.64                 & \textbf{1554.64}           & \textbf{1554.64}          & \textbf{1554.64}              & \textbf{1554.33}         & 1554.33                 & \textbf{1554.64}         & 1554.64                 & 0.00\%                   & 0.00\%                  \\
            kroA200                      & 3                     & 10691.00                & 10831.66                   & 10768.10                  & 10801.80                      & 10760.69                 & 10892.27                & \textbf{10691.03}        & 10734.69                & -0.65\%                  & -1.45\%                 \\
                                         & 5                     & 7412.12                 & 7415.54                    & 7572.32                   & 7418.87                       & 7470.78                  & 7547.11                 & 7421.01                  & 7459.46                 & -0.67\%                  & -1.16\%                 \\
                                         & 10                    & 6223.22                 & \textbf{6223.22}           & \textbf{6223.22}          & \textbf{6223.22}              & \textbf{6223.22}         & 6223.22                 & \textbf{6223.22}         & 6223.22                 & 0.00\%                   & 0.00\%                  \\
                                         & 20                    & 6223.22                 & \textbf{6223.22}           & \textbf{6223.22}          & \textbf{6223.22}              & \textbf{6223.22}         & 6223.22                 & \textbf{6223.22}         & 6223.22                 & 0.00\%                   & 0.00\%                  \\
            lin318                       & 3                     & 15663.50                & 16206.25                   & 16273.80                  & 16094.90                      & 15918.24                 & 16237.07                & 15698.61                 & 15806.81                & -1.38\%                  & -2.65\%                 \\
                                         & 5                     & 11276.80                & 11752.41                   & 11604.20                  & 11458.20                      & 11548.44                 & 11811.14                & 11289.26                 & 11372.09                & -2.24\%                  & -3.72\%                 \\
                                         & 10                    & 9731.17                 & \textbf{9731.17}           & \textbf{9731.17}          & \textbf{9731.17}              & \textbf{9731.17}         & 9731.17                 & \textbf{9731.17}         & 9731.17                 & 0.00\%                   & 0.00\%                  \\
                                         & 20                    & 9731.17                 & \textbf{9731.17}           & \textbf{9731.17}          & \textbf{9731.17}              & \textbf{9731.17}         & 9731.17                 & \textbf{9731.17}         & 9731.17                 & 0.00\%                   & 0.00\%                  \\
            att532                       & 3                     & 32223.24                & 32403.10                   & 33597.40                  & -                             & 32223.24                 & 32882.79                & \cellcolor{gray!50} \textbf{31651.32}        & 32043.34                & -1.77\%                  & -2.55\%                 \\
                                         & 5                     & 22372.68                & 22619.66                   & 23089.70                  & -                             & 22372.68                 & 22867.24                & \cellcolor{gray!50} \textbf{22007.07}        & 22266.07                & -1.63\%                  & -2.63\%                 \\
                                         & 10                    & 18059.70                & 18390.46                   & 18059.70                  & -                             & 18091.15                 & 18313.77                & \cellcolor{gray!50} \textbf{18042.64}        & 18155.17                & -0.27\%                  & -0.87\%                 \\
                                         & 20                    & 17641.16                & \textbf{17641.16}          & 17641.20                  & -                             & \textbf{17641.16}        & 17700.95                & \textbf{17641.16}        & 17641.16                & 0.00\%                   & -0.34\%                 \\
            rat783                       & 3                     & 3052.41                 & 3279.16                    & 3369.40                   & 3262.52                       & 3158.34                  & 3227.53                 & 3128.69                  & 3193.79                 & -0.94\%                  & -1.05\%                 \\
                                         & 5                     & 1961.12                 & 2092.77                    & 2127.99                   & 2066.38                       & 2024.27                  & 2077.57                 & 1996.33                  & 2037.13                 & -1.38\%                  & -1.95\%                 \\
                                         & 10                    & 1313.01                 & 1432.34                    & 1360.89                   & 1358.06                       & 1367.98                  & 1393.77                 & 1320.35                  & 1335.60                 & -3.48\%                  & -4.17\%                 \\
                                         & 20                    & 1231.69                 & 1260.88                    & \textbf{1231.69}          & \textbf{1231.69}              & \textbf{1231.69}         & 1235.00                 & \textbf{1231.69}         & 1231.84                 & 0.00\%                   & -0.26\%                 \\
            pcb1173                      & 3                     & 19569.50                & 22443.22                   & 22601.70                  & 21430.10                      & 20292.61                 & 20675.14                & 19962.93                 & 20481.85                & -1.62\%                  & -0.93\%                 \\
                                         & 5                     & 12406.60                & 14557.30                   & 14099.50                  & 13402.30                      & 12952.97                 & 13227.20                & 12812.20                 & 13130.61                & -1.09\%                  & -0.73\%                 \\
                                         & 10                    & 7623.59                 & 9222.92                    & 8160.25                   & 8120.45                       & 7864.11                  & 8000.58                 & 7681.41                  & 7827.65                 & -2.32\%                  & -2.16\%                 \\
                                         & 20                    & 6528.86                 & 7063.23                    & 6549.14                   & \textbf{6528.86}              & \textbf{6528.86}         & 6584.69                 & \textbf{6528.86}         & 6552.97                 & 0.00\%                   & -0.48\%                 \\
                                         \midrule
            Average                      & \multicolumn{1}{l}{}  & \multicolumn{1}{l}{}    & 9909.07                    & 9886.43                   & -                             & 9608.80                & 9745.51                & 9511.07                  & 9601.49                 & -0.81\%                  & -1.23\%                \\
            \bottomrule
            \end{tabular}
    \end{adjustbox}
    \caption{Results of HGA for \emph{Set IV}.
    Bold values in best columns are the values equal or better than BKS, which are the best-known solutions in the literature regardless of the computational time.
    The shaded values are the new best solutions achieved by HGA.}
    \label{tab:set4}

\end{table}

The set of instances in Table \ref{tab:set4} is introduced by \citet{wang2017memetic} and contains larger instances from TSPLIB. 
The table follows a similar structure to Table \ref{tab:set3} and presents the results of experiments conducted with HGA and baseline algorithms.
The baseline algorithms used in this set are MASVND \citep{wang2017memetic}, ES \citep{karabulut2021modeling}, HSNR \citep{he2022hybrid} and ITSHA \citep{zheng2022effective}. 
Similar to our algorithm, all algorithms stop after $n/5$ seconds, except for HSNR, which has a cutoff time of $(1.36 \times n)/5$. 
HSNR and MA were also tested on this set with a cutoff time of $2.4 \times n$ seconds, and a separate experiment was conducted to compare HGA, HSNR, ITSHA, and MA over a long period of time in section \ref{subsec:long}.
In this instance set, each instance is solved $20$ times by all algorithms, including HGA. 
To the best of our knowledge, the reported BKS column represents the best-known solutions to these problems.
Table \ref{tab:set4} shows that HGA achieves equal or better results in $22$ out of $24$ instances compared to the baseline algorithms. 
Additionally, it improves the best results (within $n/5$ seconds) for $13$ instances in this dataset. 
On average, HGA outperforms the other algorithms in $18$ instances and ties in the remaining instances out of the $24$.
Based on the current literature reports for a cutoff time of $n/5$ seconds, ITSHA has the best results on this dataset. 
However, HGA's best and average results are $0.81\%$ and $1.23\%$ lower, respectively, than ITSHA's best and average results. 
Moreover, HGA's average performance is either equal to or superior to ITSHA's average in all instances of \emph{Set IV}. 
Thus, based on the current literature, HGA has achieved the best performance for a cutoff time of $n/5$ seconds.

\subsection{Comparison of the results for a cuttoff time of $(n/100)\times 4$ minutes} \label{subsec:long}

\begin{table}[]
    \begin{adjustbox}{max width=\textwidth}
        \begin{tabular}{lrrrrrrrrrrrr}
            \toprule
                \multicolumn{1}{c}{}         & \multicolumn{1}{c}{}  & \multicolumn{1}{c}{}    & \multicolumn{2}{c}{HSNR}                           & \multicolumn{2}{c}{ITSHA}                       & \multicolumn{2}{c}{MA}                         & \multicolumn{2}{c}{HGA}                            & \multicolumn{2}{c}{Gap over MA}                    \\
                \cmidrule(lr){4-5}   \cmidrule(lr){6-7}  \cmidrule(lr){8-9}  \cmidrule(lr){10-11} \cmidrule(lr){12-13}   
                \multicolumn{1}{c}{Instance} & \multicolumn{1}{c}{$m$} & \multicolumn{1}{c}{BKS} & \multicolumn{1}{c}{best} & \multicolumn{1}{c}{avg} & \multicolumn{1}{c}{best} & \multicolumn{1}{c}{avg} & \multicolumn{1}{c}{best} & \multicolumn{1}{c}{avg} & \multicolumn{1}{c}{best} & \multicolumn{1}{c}{avg} & \multicolumn{1}{c}{best} & \multicolumn{1}{c}{avg} \\
                \midrule
                rand-100                     & 3                     & 3031.95                 & \textbf{3031.95}         & 3032.67                 & \textbf{3031.95}         & 3033.65                 & \textbf{3031.95}         & 3031.95                 & \textbf{3031.95}         & 3031.95                 & 0.00                     & 0.00                    \\
                                             & 5                     & 2409.63                 & 2411.68                  & 2415.00                 & 2411.68                  & 2414.65                 & \textbf{2409.63}         & 2409.65                 & \textbf{2409.63}         & 2409.63                 & 0.00                     & 0.00                    \\
                                             & 10                    & 2299.16                 & \textbf{2299.16}         & 2299.16                 & \textbf{2299.16}         & 2299.16                 & \textbf{2299.16}         & 2299.16                 & \textbf{2299.16}         & 2299.16                 & 0.00                     & 0.00                    \\
                                             & 20                    & 2299.16                 & \textbf{2299.16}         & 2299.16                 & \textbf{2299.16}         & 2299.16                 & \textbf{2299.16}         & 2299.16                 & \textbf{2299.16}         & 2299.16                 & 0.00                     & 0.00                    \\
                ch150                        & 3                     & 2401.63                 & 2407.34                  & 2435.49                 & 2407.34                  & 2416.87                 & \textbf{2401.63}         & 2401.86                 & \textbf{2401.63}         & 2401.92                 & 0.00                     & 0.00                    \\
                                             & 5                     & 1741.13                 & 1741.71                  & 1743.48                 & \textbf{1741.13}         & 1752.06                 & \textbf{1741.13}         & 1741.13                 & \textbf{1741.13}         & 1741.13                 & 0.00                     & 0.00                    \\
                                             & 10                    & 1554.64                 & \textbf{1554.64}         & 1554.64                 & \textbf{1554.64}         & 1554.64                 & \textbf{1554.64}         & 1554.76                 & \textbf{1554.64}         & 1554.64                 & 0.00                     & -0.01                   \\
                                             & 20                    & 1554.64                 & \textbf{1554.64}         & 1554.64                 & \textbf{1554.64}         & 1554.64                 & \textbf{1554.64}         & 1554.64                 & \textbf{1554.64}         & 1554.64                 & 0.00                     & 0.00                    \\
                kroA200                      & 3                     & 10691.00                & 10748.10                 & 10987.69                & 10700.57                 & 10819.85                & \textbf{10691.00}        & 10691.41                & \textbf{10691.03}        & 10700.06                & 0.00                     & 0.08                    \\
                                             & 5                     & 7412.12                 & 7418.87                  & 7494.44                 & 7449.22                  & 7513.67                 & \textbf{7412.12}         & 7414.21                 & 7418.87                  & 7420.23                 & 0.09                     & 0.08                    \\
                                             & 10                    & 6223.22                 & \textbf{6223.22}         & 6223.22                 & \textbf{6223.22}         & 6223.22                 & \textbf{6223.22}         & 6249.10                 & \textbf{6223.22}         & 6223.22                 & 0.00                     & -0.41                   \\
                                             & 20                    & 6223.22                 & \textbf{6223.22}         & 6223.22                 & \textbf{6223.22}         & 6223.22                 & \textbf{6223.22}         & 6223.22                 & \textbf{6223.22}         & 6223.22                 & 0.00                     & 0.00                    \\
                lin318                       & 3                     & 15663.50                & 15902.50                 & 16207.05                & 15930.04                 & 16088.56                & \textbf{15663.50}        & 15699.92                & \textbf{15663.54}        & 15714.31                & 0.00                     & 0.09                    \\
                                             & 5                     & 11276.80                & 11295.20                 & 11596.35                & 11430.65                 & 11601.67                & \textbf{11276.80}        & 11291.59                & 11288.52                 & 11297.35                & 0.10                     & 0.05                    \\
                                             & 10                    & 9731.17                 & \textbf{9731.17}         & 9731.17                 & \textbf{9731.17}         & 9731.17                 & \textbf{9731.17}         & 9731.17                 & \textbf{9731.17}         & 9731.17                 & 0.00                     & 0.00                    \\
                                             & 20                    & 9731.17                 & \textbf{9731.17}         & 9731.17                 & \textbf{9731.17}         & 9731.17                 & \textbf{9731.17}         & 9731.17                 & \textbf{9731.17}         & 9731.17                 & 0.00                     & 0.00                    \\
                att532                       & 3                     & 9966.00                 & 10231.00                 & 10565.30                & 10158.00                 & 10344.50                & \textbf{9966.00}         & 10064.00                & 9973.00                  & 10038.05                & 0.07                     & -0.26                   \\
                                             & 5                     & 6986.00                 & 7067.00                  & 7334.00                 & 7067.00                  & 7156.80                 & 6986.00                  & 7070.95                 & \cellcolor{gray!50} \textbf{6950.00}         & 7003.80                 & -0.52                    & -0.95                   \\
                                             & 10                    & 5709.00                 &  \textbf{5709.00}        & 5738.90                 & 5731.00                  & 5787.50                 & 5770.00                  & 5796.75                 & \cellcolor{gray!50} \textbf{5698.00}         & 5716.75                 & -1.25                    & -1.38                   \\
                                             & 20                    & 5580.00                 & \textbf{5580.00}         & 5580.00                 & 5583.00                  & 5601.75                 & \textbf{5580.00}         & 5589.35                 & \textbf{5580.00}         & 5580.00                 & 0.00                     & -0.17                   \\
                rat783                       & 3                     & 3052.41                 & 3187.90                  & 3237.29                 & 3131.99                  & 3180.79                 & \textbf{3052.41}         & 3083.52                 & 3086.70                  & 3106.39                 & 1.12                     & 0.74                    \\
                                             & 5                     & 1961.12                 & 2006.46                  & 2044.32                 & 2018.44                  & 2058.65                 & 1961.12                  & 1989.68                 & \cellcolor{gray!50} \textbf{1959.16}         & 1981.51                 & -0.10                    & -0.41                   \\
                                             & 10                    & 1313.01                 & 1334.76                  & 1345.88                 & 1357.65                  & 1381.69                 & 1313.01                  & 1325.54                 & \cellcolor{gray!50} \textbf{1304.70}         & 1321.33                 & -0.63                    & -0.32                   \\
                                             & 20                    & 1231.69                 & \textbf{1231.69}         & 1231.69                 & \textbf{1231.69}         & 1231.84                 & \textbf{1231.69}         & 1235.37                 & \textbf{1231.69}         & 1231.69                 & 0.00                     & -0.30                   \\
                pcb1173                      & 3                     & 19569.50                & 20813.80                 & 21144.92                & 20288.75                 & 21144.92                & \textbf{19569.50}        & 19858.77                & 19724.24                 & 20016.94                & 0.79                     & 0.80                    \\
                                             & 5                     & 12406.60                & 13032.30                 & 13216.99                & 12816.55                 & 13216.99                & \textbf{12406.60}        & 12636.49                & 12517.65                 & 12655.64                & 0.90                     & 0.15                    \\
                                             & 10                    & 7623.59                 & 7758.26                  & 7897.20                 & 7801.18                  & 7910.09                 & 7623.59                  & 7745.00                 & \cellcolor{gray!50} \textbf{7512.43}         & 7617.21                 & -1.46                    & -1.65                   \\
                                             & 20                    & 6528.86                 & \textbf{6528.86}         & 6528.86                 & \textbf{6528.86}         & 6534.75                 & \textbf{6528.86}         & 6548.87                 & \textbf{6528.86}         & 6528.86                 & 0.00                     & -0.31                   \\
                                             \midrule
                Average                      &                       & 6922.17                 & 7042.20                  & 7139.50                 & 7016.30                  & 7115.04                 & 6924.71                  & 6967.85                 & 6928.72                  & 6962.13                 & -0.03                    & -0.15                  \\
                \bottomrule
            \end{tabular}
    \end{adjustbox}
    \caption{Results of HGA with a cutoff time of $(n/100)\times 4$ minutes.
    Bold values in best columns are the values equal to or better than BKS, which are the best-known solutions in the literature regardless of the computational time.
    The shaded values are the new best solutions achieved by HGA.}
    \label{tab:long}
\end{table}

In order to compare HGA with MA \citep{he2023memetic}, we conduct experiments on \emph{Set III}, \emph{Set IV}, and an additional instance called rand100. 
The cutoff time used for these experiments is $(n/100)\times 4$ minutes, as used by MA. 
The baseline algorithms in this section are HSNR \citep{he2022hybrid}, ITSHA \citep{zheng2022effective}, and MA \citep{he2023memetic}.
The reported results for ITSHA are provided by \citet{he2023memetic}, and we trust their reported results and present them as they are. 
It may be unclear to the reader why the results of the att532 instance are not consistent with Table \ref{tab:set4}. 
This discrepancy arises because both \citet{he2022hybrid} and \citet{he2023memetic} use pseudo-Euclidean distances for this instance, as specified in the TSPLIB documentation. 
Consequently, we perform the same analysis using pseudo-Euclidean distances for HGA as well.
Table \ref{tab:long} shows the results of the comparison between HGA and MA. 
On average, HGA has a slight advantage over MA, with a difference of $0.15 \%$. 
However, we cannot claim that HGA outperforms MA significantly due to the small difference. 
Among the $28$ instances, HGA finds solutions that are equal to or better than the best-known solutions in $22$ instances. 
Additionally, HGA identifies five new best solutions. 
In terms of average results across the $28$ instances, HGA outperforms MA in $10$ cases, loses in $7$ cases, and ties in $11$ cases.

\begin{table}[]
    \begin{adjustbox}{max width=\textwidth}
        \begin{tabular}{rr r rr rr rr}
            \toprule
            \multicolumn{2}{c}{Crossover}                 & \multicolumn{1}{c}{OX}  & \multicolumn{2}{c}{STX}                             & \multicolumn{2}{c}{OX}                              & \multicolumn{2}{c}{STX}  
                           \\
			\cmidrule(lr){1-2}
			\cmidrule(lr){3-3}
			\cmidrule(lr){4-5}
			\cmidrule(lr){6-7}
			\cmidrule(lr){8-9}
            \multicolumn{2}{c}{intersections}   & \multicolumn{1}{c}{not removed}  & \multicolumn{2}{c}{not removed}              & \multicolumn{2}{c}{removed}                             & \multicolumn{2}{c}{removed}                             \\
			\cmidrule(lr){1-2}
			\cmidrule(lr){3-3}
			\cmidrule(lr){4-5}
			\cmidrule(lr){6-7}
			\cmidrule(lr){8-9}
			$n$ & $m$ & \multicolumn{1}{c}{avg (10 runs)} & \multicolumn{1}{c}{avg} & \multicolumn{1}{c}{imprv} & \multicolumn{1}{c}{avg} & \multicolumn{1}{c}{imprv} & \multicolumn{1}{c}{avg} & \multicolumn{1}{c}{imprv} \\
            \midrule
            76                    & 2                     & 283.91                  & 283.28                  & -0.22\%                   & 281.38                  & -0.89\%                   & 281.62                  & -0.81\%                   \\
            99                    & 3                     & 520.44                  & 520.74                  & 0.06\%                    & 517.72                  & -0.52\%                   & 517.81                  & -0.51\%                   \\
            99                    & 7                     & 438.01                  & 437.88                  & -0.03\%                   & 437.58                  & -0.10\%                   & 437.31                  & -0.16\%                   \\
            150                   & 3                     & 13287.01                & 13154.68                & -1.00\%                   & 13230.87                & -0.42\%                   & 13093.48                & -1.46\%                   \\
            150                   & 5                     & 8679.41                 & 8591.76                 & -1.01\%                   & 8594.86                 & -0.97\%                   & 8486.81                 & -2.22\%                   \\
            150                   & 3                     & 2468.89                 & 2440.98                 & -1.13\%                   & 2423.64                 & -1.83\%                   & 2405.79                 & -2.56\%                   \\
            200                   & 3                     & 11078.12                & 10822.11                & -2.31\%                   & 10961.14                & -1.06\%                   & 10734.69                & -3.10\%                   \\
            200                   & 5                     & 7641.64                 & 7545.3                 0 & -1.26\%                   & 7551.27                 & -1.18\%                   & 7459.46                 & -2.38\%                   \\
            318                   & 3                     & 16510.62                & 15981.36                & -3.21\%                   & 16276.37                & -1.42\%                   & 15806.81                & -4.26\%                   \\
            318                   & 5                     & 11854.43                & 11643.22                & -1.78\%                   & 11713.21                & -1.19\%                   & 11372.09                & -4.07\%                   \\
            783                   & 10                    & 1482.70                 & 1399.26                 & -5.63\%                   & 1394.32                 & -5.96\%                   & 1335.60                  & -9.92\%                   \\
            783                   & 20                    & 1255.48                 & 1245.00                 & -0.83\%                   & 1239.74                 & -1.25\%                   & 1231.84                 & -1.88\%                   \\
            1173                  & 10                    & 8687.84                 & 8300.38                 & -4.46\%                   & 8347.27                 & -3.92\%                   & 7827.65                 & -9.90\%                   \\
            1173                  & 20                    & 6714.53                & 6697.22                 & -0.26\%                   & 6651.21                 & -0.94\%                   & 6552.97                 & -2.14\%                   \\
            \midrule
            \multicolumn{2}{c}{Average}                   & 6478.63                 & 6361.66                 & -1.65\%                   & 6401.47                 & -1.55\%                   & 6253.14                 & -3.26\%                  \\
            \bottomrule
        \end{tabular}

\end{adjustbox}
            \caption{Contribution of STX crossover and removing the intersections.}

    \label{tab:comparison}
\end{table}

\subsection{Effectiveness of STX and removing intersections} \label{subsec:effect}
In this section, we conduct an experimental analysis to evaluate the effectiveness of our proposed STX crossover technique, as well as the impact of removing intersections in the context of our problem. 
A number of instances are selected from various benchmark sets and range in size from 76 to 1173 with different numbers of salesmen.
To assess the effectiveness of the STX crossover, we compare its performance with the order-based crossover (OX) method, which is described in \citet{larranaga1999genetic}. 
Additionally, we examine the impact of removing intersections by solving the instances both with and without this operation.
Table \ref{tab:comparison} presents the results obtained from different combinations of crossover techniques and the presence or absence of intersection removal. 
To evaluate the effectiveness of STX, we focus on comparing the columns where the status of intersection removal is the same. 
When intersections are not removed, employing STX results in an average improvement of $1.65\%$ (as observed by comparing the first and second columns). 
However, when intersections are removed, utilizing STX leads to a greater improvement of $2.31\%$ (as shown in the third and fourth columns).
It appears that they may have a synergistic effect. 
Likewise, we investigate the impact of removing intersections by fixing the crossover function. 
In this case, we observe an improvement of $1.55\%$ on average by comparing the third and first columns, where OX is used. 
Furthermore, removing intersections contributes to an average improvement of $1.71\%$ when STX is employed (as seen in the fourth and second columns).

\section{Conclusions and future directions} \label{sec:conclusion}
This paper presented a hybrid genetic algorithm (HGA) for solving the multiple traveling salesman problem (mTSP) with a min-max objective function. 
The HGA utilized TSP sequences to represent solutions and employed a dynamic programming-based algorithm called \SPLIT{} to determine the optimal mTSP solution for each sequence.
To generate tours for the offspring, a novel crossover algorithm called STX was designed, which combines similar tours from the parent solutions. 
The HGA further enhanced solutions using different neighborhoods as local searches. 
Moreover, the paper introduced a novel approach to detect and remove intersections between different tours, which proved to be highly effective for mTSPs with min-max objectives.
The HGA was evaluated on four different datasets, comprising a total of $89$ instances. 
In $76$ out of the $89$ problems, HGA's best solution either matched or surpassed the existing best-known solutions. 
Additionally, HGA improved the best-known solutions in $21$ instances. 
Furthermore, HGA demonstrated equal or superior average performance in $78$ out of the $89$ problems compared to the baseline algorithm.
Overall, the paper presented a hybrid genetic algorithm that leverages TSP sequences, dynamic programming, novel crossover, local search, and intersection detection techniques to solve the mTSP with a min-max objective. 
The experimental results indicated the effectiveness of the proposed approach in achieving high-quality solutions and outperforming the baseline algorithms in terms of both best and average performance.

In addition to the current findings, there are several promising avenues for future research in the domain of the min-max multiple traveling salesman problem (mTSP). 
One potential direction is to extend the proposed hybrid genetic algorithm (HGA) to handle the multi-depot mTSP. 
This would require modifications to the \SPLIT{} algorithm and the crossover functions to accommodate multiple depots. 
Furthermore, considering the emerging field of drone technology, there is an opportunity to explore the integration of drones as agents in the mTSP. 
This would involve addressing challenges such as limited flight time. 

\section*{Acknowledgment}

This material is based upon work supported by the National Science Foundation under Grant No. 2032458 and funded by the Ministry of Science and ICT of Korea via the National Research Foundation of Korea (NRF) under Grant No. 2022M3J6A1063021.

\bibliographystyle{ormsv080-ck}
\bibliography{ref}

\end{document}